%% file: tai_template.tex
\documentclass[journal]{IEEEtai}

\usepackage[colorlinks,urlcolor=blue,linkcolor=blue,citecolor=blue]{hyperref}

\usepackage{color,array}

\usepackage{graphicx}

\usepackage{microtype}
\usepackage{graphicx}
\usepackage{multirow}
\usepackage{arydshln}
\let\labelindent\relax
\usepackage{enumitem,colortbl}
\usepackage{times}
\usepackage{latexsym}
\usepackage{subfig}
\usepackage{amsmath}
\usepackage{amssymb}
\usepackage{soul}
\usepackage[T1]{fontenc}
\usepackage[utf8]{inputenc}

\definecolor{LightCyan}{rgb}{0.88,1,1}
\newcommand{\dataset}{\texttt{SPICE}}
\newcommand{\model}{\texttt{SPOT}}

\newcommand{\citet}{\cite}
\newcommand{\citep}{\cite}

\begin{document}

\title{Speaker Profiling in Multiparty Conversations}

\author{Shivani Kumar, Rishabh Gupta, Md Shad Akhtar, Tanmoy Chakraborty
\thanks{Submitted for review on 10th April 2023. This work was supported by ihub-Anubhuti-iiitd Foundation, set up under the NM-ICPS scheme of the DST.}
\thanks{Shivani Kumar is with Indraprastha Institute of Information Technology Delhi, India. (e-mail: shivaniku@iiitd.ac.in).}
\thanks{Rishabh Gupta was with Indraprastha Institute of Information Technology Delhi, India. (e-mail: rishabh19089@iiitd.ac.in).}
\thanks{Md Shad Akhtar is with Indraprastha Institute of Information Technology Delhi, India. (e-mail: shad.akhtar@iiitd.ac.in)}
\thanks{Tanmoy Chakraborty is with Indian Institute of Technology Delhi, India. (e-mail: tanchak@iitd.ac.in)}
\thanks{This paragraph will include the Associate Editor who handled your paper.}}

\markboth{Journal of IEEE Transactions on Artificial Intelligence, Vol. 00, No. 0, Month 2022}
{Shivani Kumar \MakeLowercase{\textit{et al.}}: Speaker Profiling in Multiparty Conversations}

\maketitle

\begin{abstract}
\input{1_abstract}
\end{abstract}

\begin{IEEEImpStatement}
Incorporating individual variability in response generation is crucial as people react differently to comments. Therefore, identifying a person's personality characteristics, including their preferences and aversions, is necessary to determine their potential response to a specific remark. This paper focuses on extracting such profiles for individuals participating in a conversation, which can then be utilized by response selection methods to generate more effective replies.
\end{IEEEImpStatement}

\begin{IEEEkeywords}
Speaker profiling, personalisation, dialogue systems, dialogue understanding, natural language processing.
\end{IEEEkeywords}

\vspace{-3mm}
\section{Introduction}
    \label{sec:introduction}
    \input{2_introduction}
\section{Related Work}
    \label{sec:related_work}
    \input{4_related_work}
\section{Problem Statement}
    \label{sec:problem_statement}
    \input{3_problem_statement}
\section{Dataset}
    \label{sec:dataset}
    \input{5_dataset_coling.tex}
    \input{tab-dataset-stats}
\section{Proposed Methodology}
    \label{sec:methodology}
    \input{6_methodology_coling.tex}
\section{Experiments and Results}
    \label{sec:results}
    \input{7_results_new}
\section{Error Analysis}
    \label{sec:result_analysis}
    \input{8_result_analysis_new}
\section{Conclusion}
    \label{sec:conclusion}
    \input{9_conclusion_new}

\bibliographystyle{IEEEtran}
\bibliography{tai_template}

\begin{IEEEbiography}[{\includegraphics[width=1in,height=1.25in,clip,keepaspectratio]{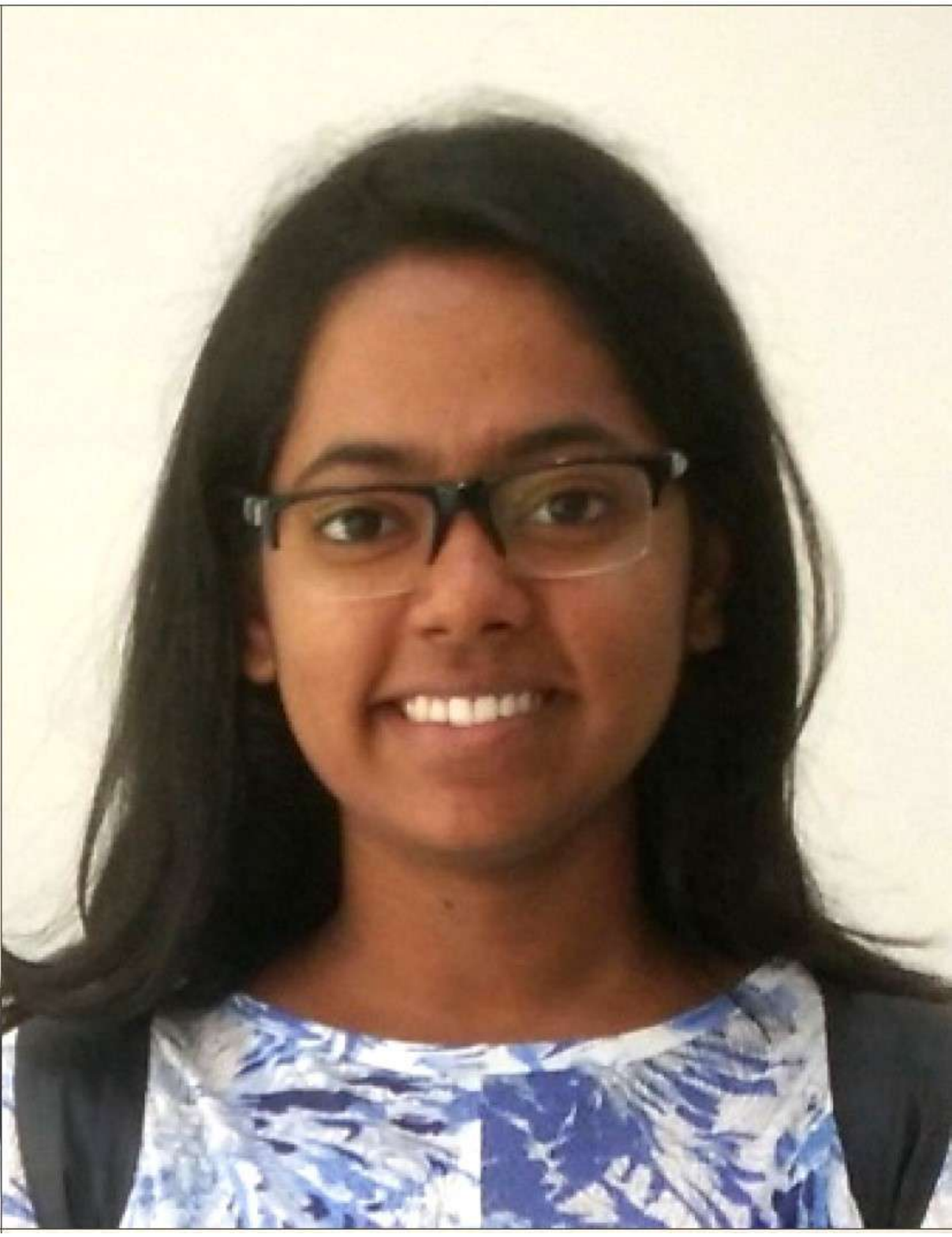}}]{Shivani Kumar}{\space} is a PhD scholar at Indraprastha Institute of Information Technology Delhi (IIIT Delhi), India. She holds a Senior Research Fellowship and works in the domain of Natural Language Processing, primarily in the area of understanding and explaining various affects, like emotions, sarcasm, and humour, in conversational data. 
\end{IEEEbiography}

\begin{IEEEbiography}[{\includegraphics[width=1in,height=1.25in,clip,keepaspectratio]{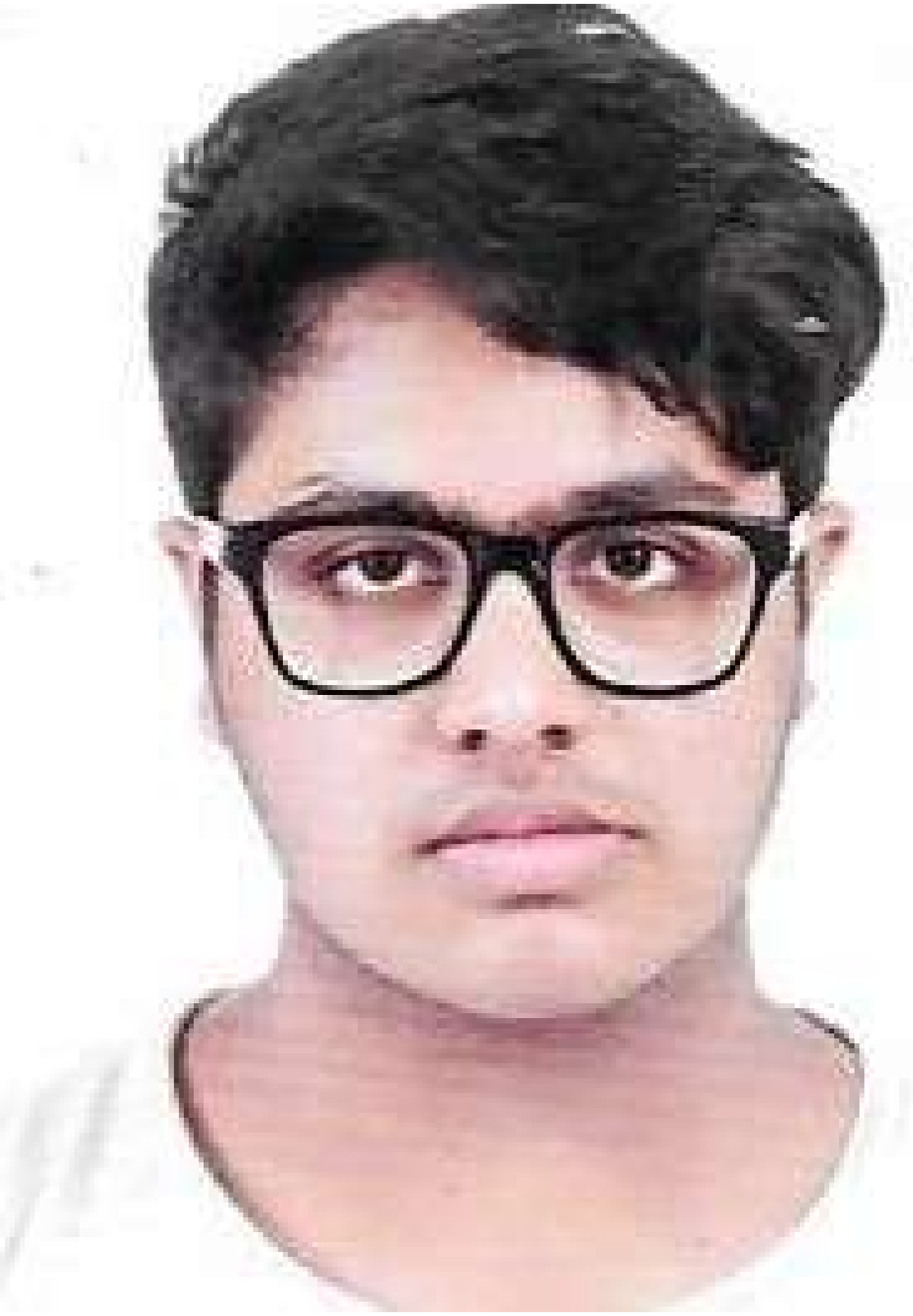}}]{Rishabh Gupta}{\space} was a bachelors student at Indraprastha Institute of Information Technology Delhi (IIIT Delhi), India for the duration of this work. His research interest lies in Natural Language Processing and its sub-fields. 
\end{IEEEbiography}

\begin{IEEEbiography}[{\includegraphics[width=1in,height=1.25in,clip,keepaspectratio]{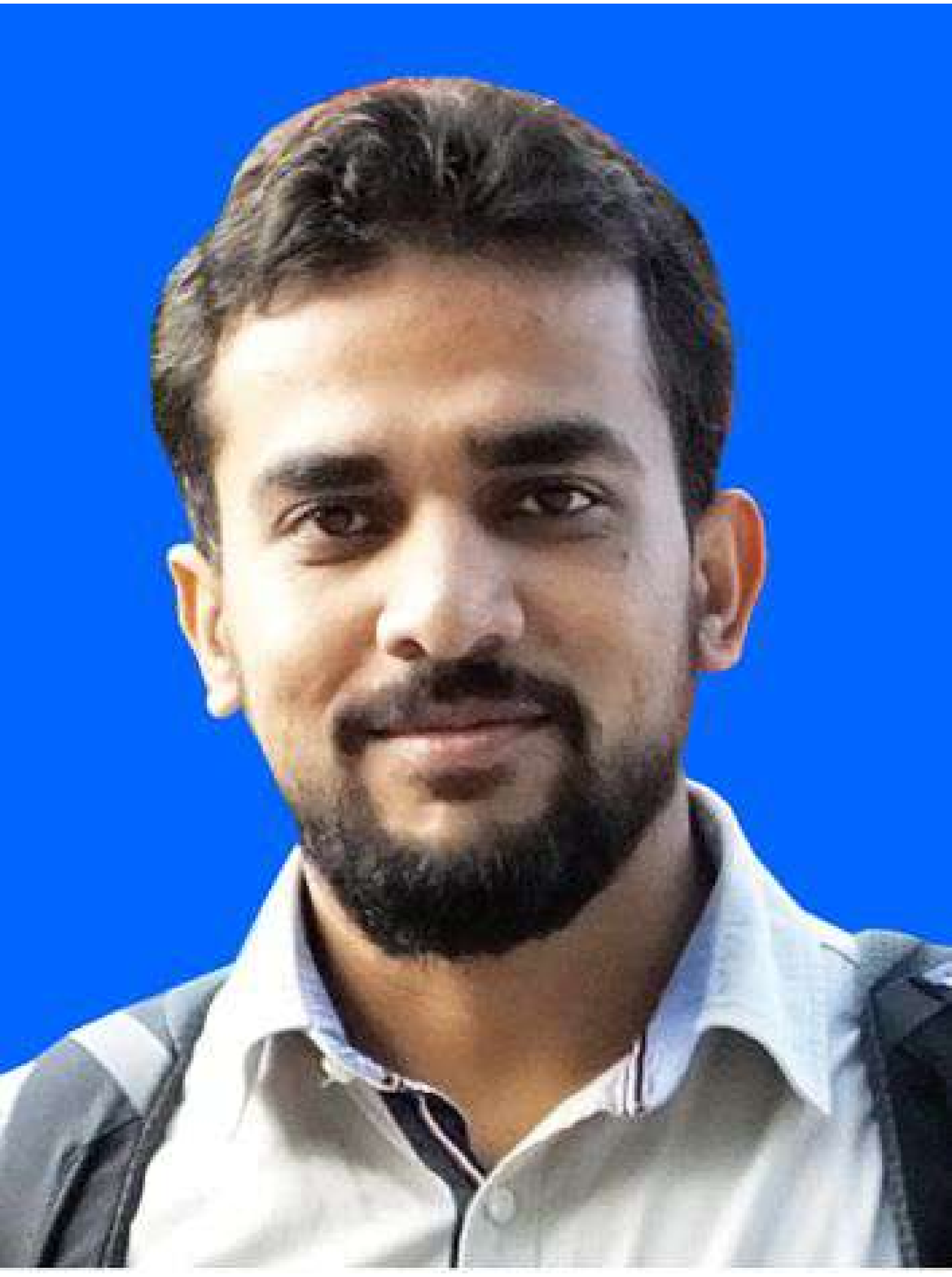}}]{Md Shad Akhtar}{\space} is currently an Assistant Professor at Indraprastha Institute of Information Technology Delhi (IIIT Delhi). His main area of research is NLP with a focus on the affective analysis. He completed his PhD from IIT Patna.
\end{IEEEbiography}

\begin{IEEEbiography}[{\includegraphics[width=1in,height=1.25in,clip,keepaspectratio]{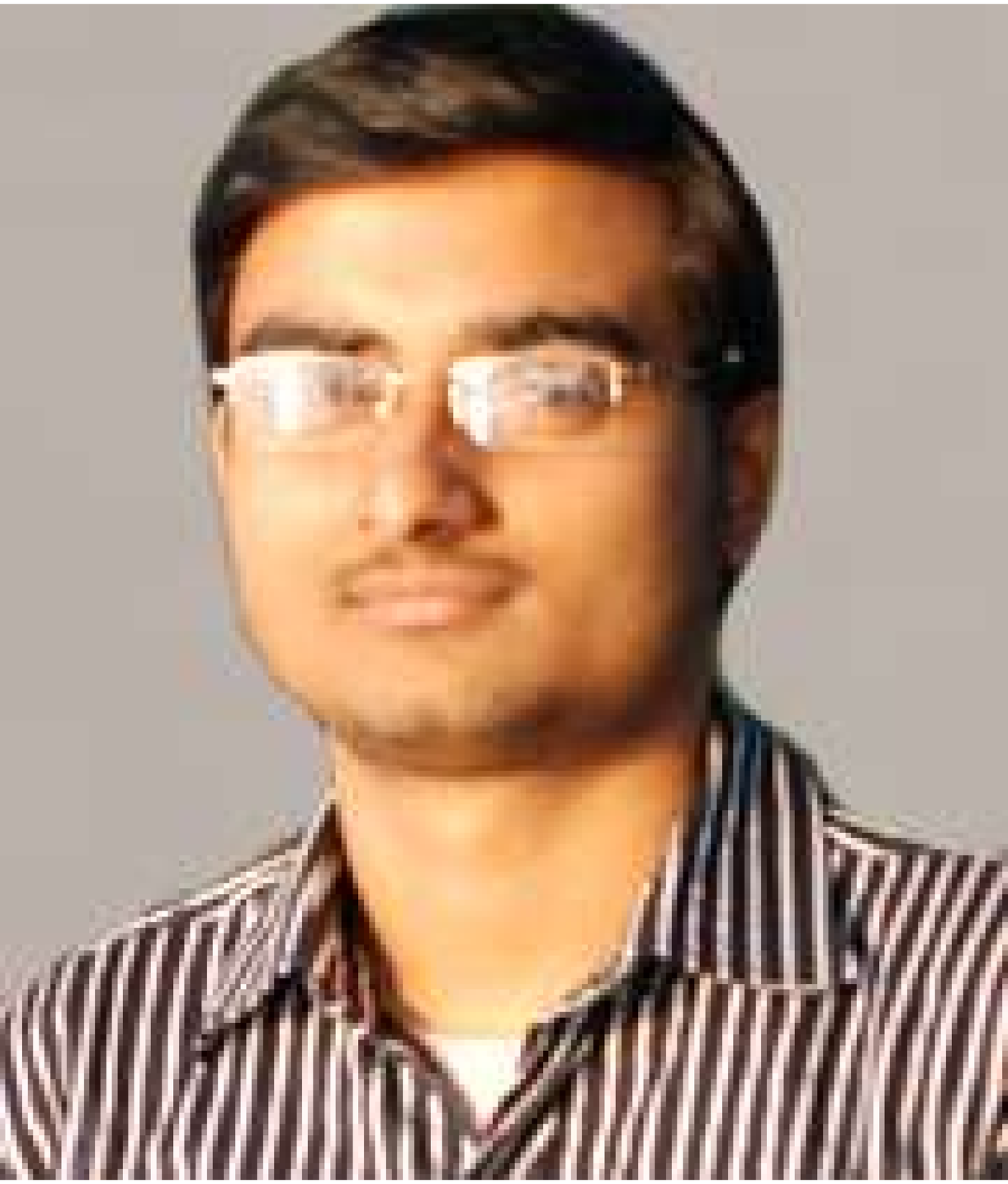}}]{Tanmoy Chakraborty}{\space} is an Associate Professor in the Dept. of Electrical Engineering, Indian Institute of Technology Delhi, India since September 2022. Before joining IIT Delhi, he served as an Associate Professor in the Dept of CSE, Indraprastha Institute of Information Technology Delhi (IIIT Delhi), India. He completed his postdoctoral research from University of Maryland, College Park after obtaining his PhD from the Dept. of CSE, IIT Kharagpur, India as a Google PhD scholar. His broad research interests include Natural Language Processing, Graph Neural Networks, and Social Computing.
\end{IEEEbiography}

\end{document}

%% file: 1_abstract.tex
In conversational settings, individuals exhibit unique behaviors, rendering a one-size-fits-all approach insufficient for generating responses by dialogue agents. Although past studies have aimed to create personalized dialogue agents using speaker persona information, they have relied on the assumption that the speaker's persona is already provided. However, this assumption is not always valid, especially when it comes to chatbots utilized in industries like banking, hotel reservations, and airline bookings. This research paper aims to fill this gap by exploring the task of Speaker Profiling in Conversations (SPC). The primary objective of SPC is to produce a summary of persona characteristics for each individual speaker present in a dialogue. To accomplish this, we have divided the task into three subtasks: persona discovery, persona-type identification, and persona-value extraction. Given a dialogue, the first subtask aims to identify all utterances that contain persona information. Subsequently, the second task evaluates these utterances to identify the type of persona information they contain, while the third subtask identifies the specific persona values for each identified type. To address the task of SPC, we have curated a new dataset named \dataset, which comes with specific labels. We have evaluated various baselines on this dataset and benchmarked it with a new neural model, \model, which we introduce in this paper. Furthermore, we present a comprehensive analysis of \model, examining the limitations of individual modules both quantitatively and qualitatively.

%% file: 2_introduction.tex
Understanding natural language inputs is essential to process them effectively \cite{schank1972conceptual,pruksachatkun2020intermediate}. While a considerable amount of work has been done on standalone texts such as tweets \cite{zhou2013sentiment,badjatiya2017deep}, recent efforts have focused on contextual conversation data. In conversations, speakers need to understand each other to communicate effectively, and much research has been conducted in fields such as emotion analysis \cite{poria2019emotion,jiao2020real,shen2020dialogxl}, intent identification \cite{larson2019evaluation,gangadharaiah2019joint}, and dialogue act detection \cite{qin2020co,liu2017using}, among others. However, with the proliferation of dialogue agents in various fields, it is no longer sufficient to merely understand the conversation; generating valid responses has become a necessity.
As a result, numerous studies have explored the field of dialogue generation \cite{wu2018dialog,saleh2020hierarchical}, with the ability to engage participants being one of the most important measures of success \cite{wu2018dialog,saleh2020hierarchical}. To this end, researchers have investigated empathetic \cite{lin2020caire,shin2019happybot,rashkin2018towards} and stylistic dialogue generation \cite{su2020stylistic,akama2017generating,danescu2011chameleons}.
While an empathetic and eloquent agent can make a dialogue system more interesting, it still lacks the capability to personalize responses.
To this end, various studies have been conducted to develop a personalised dialogue generation system which is provided with the users' persona as input \cite{zhang2018personalizing,weston2018retrieve,roller2020recipes,dinan2018wizard,chen2020listener}.
A persona is a collection of statements describing different aspects of an individual, such as ``Jack likes to play cricket" \cite{zhang2018personalizing}. By providing persona information as input to dialogue agents, the generated responses become more intuitive and engaging for the user \cite{zhang2018personalizing,weston2018retrieve,roller2020recipes,dinan2018wizard}. However, the existing studies in this domain assume that the speaker's persona is already provided to the model in advance, which is often not the case. For instance, a hotel reservation chatbot cannot be supplied with the persona information of the customer beforehand.

\begin{figure}[t]
    \centering
    \subfloat[Explicit Persona \label{fig:data_eg1}]{
    \includegraphics[width=\columnwidth]{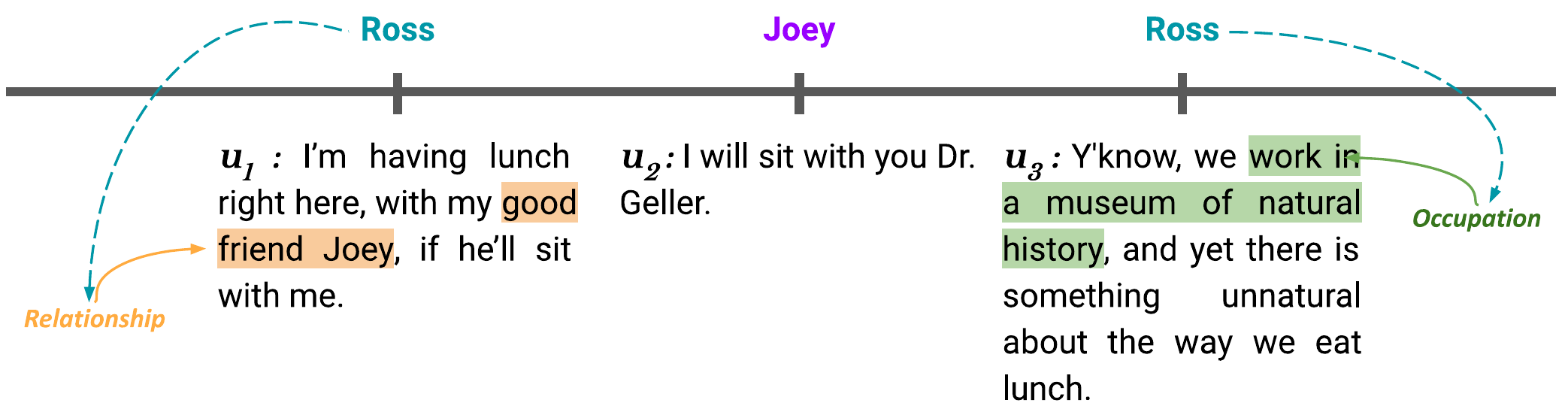}}\\
    \subfloat[Implicit Persona \label{fig:data_eg2}]{
    \includegraphics[width=\columnwidth]{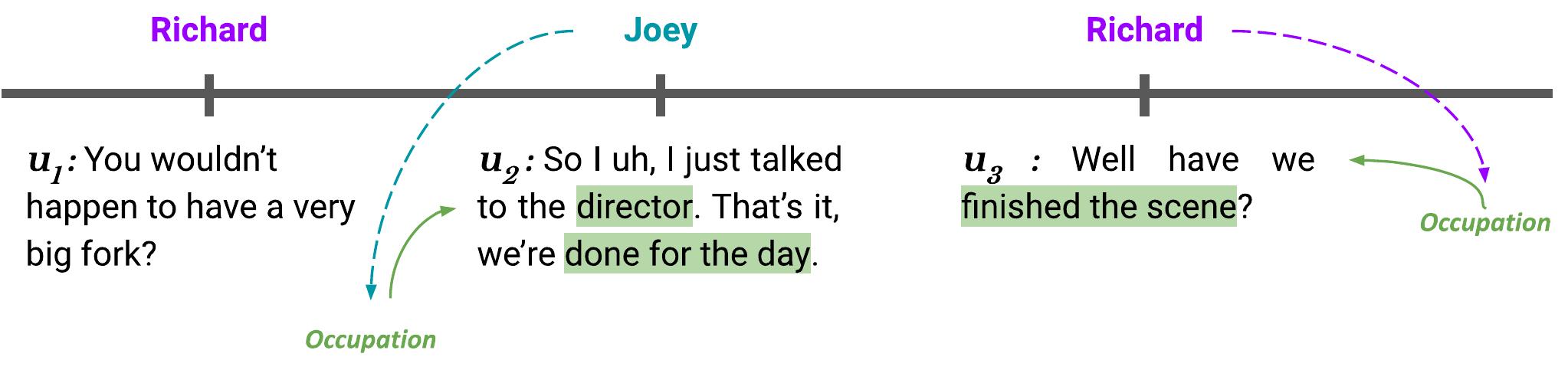}}
    \caption{Persona information in dialogues.}
    \label{fig:data_eg}
    \vspace{-5mm}
\end{figure}

To tackle the issue of persona information unavailability in chatbots, we delve into the task of \textbf{Speaker Profiling in Conversations (SPC)}. SPC aims to generate comprehensive profiles of all speakers involved in a conversation by identifying various speaker-centric attributes such as traits, likes, and relationships. To illustrate SPC, we present two dialogues in Figure \ref{fig:data_eg}. In the first dialogue (Figure \ref{fig:data_eg1}), Ross characterizes Joey as his ``good friend," indicating a friendly relationship between them. Ross also confirms his occupation as working in a natural history museum in his third utterance ($u_3$). While identifying explicit relationship and occupation information is easy in curated dialogues, natural conversations often have implicit information that is challenging to identify. The second dialogue in Figure \ref{fig:data_eg2} demonstrates how extracting persona information about the field of occupation can be non-trivial, requiring specialized knowledge to link phrases such as "director" and "scene" to a profession in the movie industry. In this paper, we address both explicit and implicit forms of persona identification.




The task of speaker profiling can be broken down into three subtasks: \textit{persona discovery}, \textit{persona-type identification}, and \textit{persona-value extraction}. The first subtask involves identifying which utterances in the conversation contain persona information. The second subtask involves determining the type of persona information that each identified utterance contains. Finally, the last subtask involves extracting the exact value for each identified persona type. In order to facilitate research in this area, we introduce \dataset\footnote{\dataset: \textbf{S}peaker \textbf{P}rofiling \textbf{I}n \textbf{C}onv\textbf{E}rsation}, a new dataset containing multi-party conversations along with annotated labels for all three subtasks. Additionally, we propose \model\footnote{\model: \textbf{S}peaker \textbf{P}r\textbf{O}filing using \textbf{T}ransformers}, a neural method based on GRU and Transformers, which is able to capture both word-level and dialogue-level context for persona discovery, and also incorporates speaker-level semantics and boundary loss \cite{zhang2021deep} for persona-type identification. We evaluate \model\ and four baseline methods in both standalone and pipeline settings and find that \model\ outperforms the baselines for both subtasks. We also conduct an analysis of the individual components of \model\ to better understand its strengths and limitations.

In a nutshell, our contributions are four-fold:
\begin{enumerate}[leftmargin=*]
    \item We explore the problem of \textbf{Speaker Profiling in Conversations} from a new angle, where given a dialogue as input, the task is to extract the speaker-centric personality information of all speakers present in the dialogue.
    \item We curate \dataset, a {\bf multi-party conversation dataset} with human annotated SPC labels.
    \item We benchmark it with a Transformers and GRU based {\bf novel model}, \model\ that we develop for this task.
    \item We perform a {\bf comparative analysis} of our model with several baselines and establish the superiority of {\model}.
\end{enumerate}

\textbf{Reproducibility:} The source code for \model\ and the \dataset\ dataset (partial) along with the execution instructions can be found here: \href{https://drive.google.com/drive/folders/1cw1iglf3BueUUbGBNKhKW1UZo22uTCPS?usp=sharing}{bit.ly/3o4sWIU}.


%% file: 4_related_work.tex
Various studies have focused on natural language understanding in conversations, including intent classification \cite{larson2019evaluation,gangadharaiah2019joint}, dialogue act recognition \cite{qin2020co,liu2017using}, and emotion analysis \cite{poria2019emotion,jiao2020real,shen2020dialogxl}. The primary aim of comprehending conversations is to develop more engaging dialogues. One way to achieve this is by catering to the interests of the users. This can be illustrated by the following example: Suppose Andrew wants to go on a date with Lisa, and he already knows her likes and dislikes. With this knowledge, Andrew can not only arrange an outstanding date but also engage in captivating conversations with Lisa. Similarly, for online dialogue agents, such additional information can enhance dialogue generation.

\textbf{Personalised dialogue systems.} It has been widely recognized that personalization improves the performance of dialogue systems \cite{weston2018retrieve,roller2020recipes,dinan2018wizard,chen2020listener,lucas2009managing,joshi2017personalization}. While some studies have focused on using user profiles to tailor goal-oriented dialogue systems \cite{lucas2009managing,joshi2017personalization}, recent research has shifted towards chit-chat settings.
In the past, personalization in vector form has been employed, such as in \citet{li2016persona}, where distributed embeddings were learned for each Twitter user to capture individual characteristics like writing style and past experience. The authors demonstrated that incorporating these vectors improved the performance of a sequence-to-sequence model. Later, \citet{zhang2018personalizing} introduced the Persona-chat dataset, which consists of $10,907$ dyadic dialogues where speakers were assigned fictitious personas and had to converse according to them. The dataset included persona information of the speakers and baselines for the next utterance generation were provided. The authors showed that including persona information in the generation process is necessary for better results.
Following this, a few studies showed the benefit of the Persona-chat dataset for personalization while generating replies to the user \cite{weston2018retrieve,roller2020recipes,dinan2018wizard}. Although leveraging persona information for dialogue generation results in better responses, it is vital not to presume that this information is freely available.

\textbf{Persona identification in dialogues.} 
In their study, \citet{chu2018learning} tackled the challenge of character archetype assignment in discourse by utilizing the CMU Movie Summary dataset \cite{bamman2013learning}. They developed a dialogue dataset by using IMDB dialogue snippets to conduct conversational dialogues with characters from the CMU dataset. To classify characters into tropes, the authors employed a multi-level attention mechanism to capture the dialogue essence.
Meanwhile, \citet{tigunova2019listening} explored the extraction of speaker qualities from conversations using the MovieChAtt dataset, as well as additional datasets such as Persona-chat and Reddit. However, they only considered a limited set of speaker attributes, including profession, gender, and family status, which cannot entirely represent a speaker's persona. In contrast, \citet{wu2019getting} proposed a two-stage attribute extractor to obtain user attributes as triplets of (subject, predicate, object). Nevertheless, their dataset and model were only applicable to dyadic interactions and could not be extended to multi-party scenarios.
Recently, \citet{gu2021detecting} utilized the Persona-chat dataset and introduced the task of speaker persona detection, which categorizes each speaker into one of the predefined personas. In the audio domain as well, persona identification is relatively well explored \cite{okada2015acm,gilpin2018perception}. \citet{kalluri2020automatic} attempted to estimate different physical parameters of the speaker, such as age, height, weight, and shoulder size, based on a brief duration of the speech. \citet{schilling_marsters_2015} dived into the world of forensic speaker profiling and highlighted various characteristics uncovered by audio transcript speaker profiling.
However, such characteristics are often physical, associated with the human physique.

Our study differs from others as our aim is to create a natural language summary of the personas of all speakers involved in a conversation, solely based on the input dialogue. To achieve this, we focus on five persona types that encompass all the crucial characteristics of a human persona. Additionally, our method is indifferent to the number of speakers present in the conversation. These distinctions will be elaborated on in the following sections.

\vspace{1mm}

\begin{figure}[t]
    \centering
    \includegraphics[width=\columnwidth]{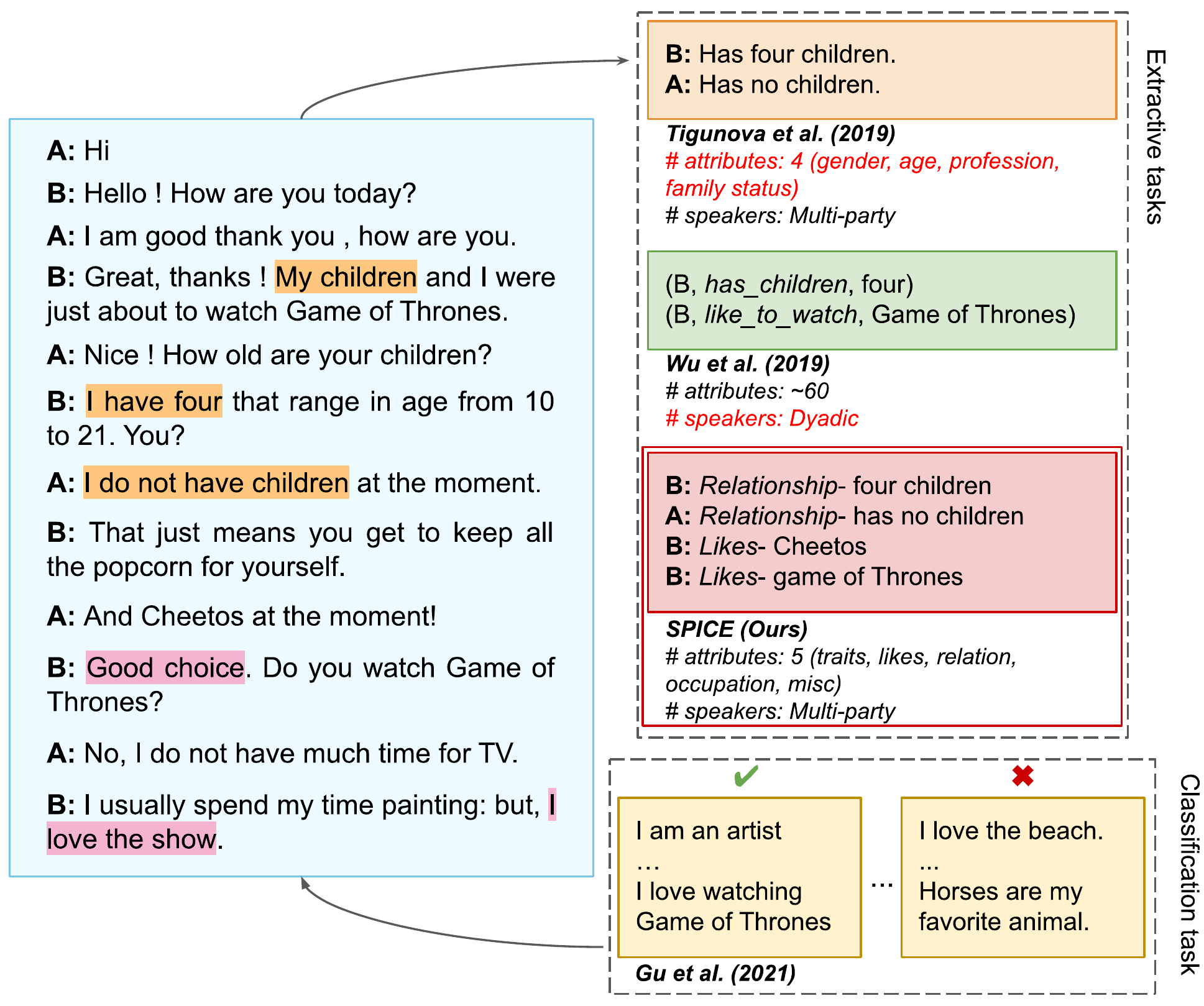}
    \caption{Difference between speaker profiling in conversation (proposed) and related works.}
    \label{fig:compare_task}
    \vspace{-5mm}
\end{figure}

\textbf{How is Our Task Different?}
\label{sec:app_compare_tasks}
Several studies have attempted to extract speaker characteristics from dialogues, as previously noted. To illustrate the differences between these studies, Figure \ref{fig:compare_task} presents a sample dialogue. \citet{tigunova2019listening} extracted four types of persona information, including profession, gender, age, and family status, while \citet{wu2019getting} extracted information in the form of triplets, both of which used heuristics to collect ground-truth labels without the use of human-level gold labels. In contrast, our solution is based on human-annotated ground-truth labelling, which ensures cleaner and less noisy data. It is important to note that while \citet{wu2019getting} is only able to handle dyadic conversations, our task can also handle multi-party scenarios. \citet{gu2021detecting} projected the task of speaker profiling as a classification task, determining the ranking among the available personalities and assigning the most appropriate identity to the dialogue. However, in this case, a speaker can only be assigned a pre-existing persona. Our proposed work, on the other hand, extracts persona on the fly.

Many datasets have been created for speaker personality-related tasks over time, starting with Persona-chat \cite{zhang2018personalizing}. This dataset contains a manually assigned persona for each speaker in a conversation, but there is no mapping between the persona and the utterances in the discussion. To address this, the Dialogue-NLI dataset was introduced by \citet{welleck2018dialogue}, which expanded Persona-chat to associate a triplet-based persona knowledge at the utterance-level in a conversation. However, this mapping was not based on human-level annotations. Our proposed research overcomes these issues by verifying human-annotated ground-truth labels for precisely defined persona-types in multi-party discussions.


%% file: 3_problem_statement.tex
The task of SPC can be conceptualized as a synthesis of three subtasks. Formally, we define the subtasks below:

\begin{itemize}[leftmargin=*,noitemsep,topsep=0pt]
    \item   \textbf{\textit{{Persona discovery}}:} Given a sequence of $n$ utterances as a dialogue, $D = \{u_1, u_2, …, u_n\}$, in a multiparty conversation where $u_i$ represents the $i^{th}$ utterance of the dialogue, we associate a binary label with $u_i$, $i \in [1,n]$, signifying whether $u_i$ contains a persona information of the speaker articulating that utterance.
    \item \noindent \textbf{\textit{{Persona-type identification}}:} Given an instance, $I_p = \{u_1, u_2, ..., u_m\}$, where $u_m$ is identified as an utterance containing persona information, we associate a label $p$, which represents what type of persona is present in the concerned utterance. Here, $p$ belongs to a set of carefully curated possible persona types $P$. Section \ref{sec:dataset} elaborates on the types of persona we consider.
    \item \noindent \textbf{\textit{Persona-value extraction}:} Given an instance, $I_v = \{u_1, u_2, ..., u_m\}$, where $u_m$ contains information about the persona type $p$, the task is to extract the exact persona value $v$ associated with $p$ from $I_v$.
\end{itemize}




%% file: 5_dataset_coling.tex
\input{tab-instigators-defn}
\input{tab-dataset-instance}

We present a new dataset, called \dataset, for speaker profiling in multi-party conversations. We extract conversations from the MELD dataset \citep{poria2018meld} and then annotate each utterance for our task. Following MELD's original train-dev-test split, we carry out three subtasks for annotation.

\begin{itemize}[leftmargin=*,noitemsep,topsep=0pt]
    \item \textit{Persona discovery:} We identify the presence of persona information in each utterance of the given dialogue by marking it as ``yes" in the \textit{persona discovery} subtask.
    \item \textit{Persona-type identification:} We associate a type of persona to each utterance marked as ``yes" in the previous phase, in the \textit{persona-type identification} subtask. After carefully analyzing each conversation in the dataset, we define five persona types - \textit{trait}, \textit{likes}, \textit{relation}, \textit{occupation}, and \textit{misc} - to capture different personality characteristics of the speakers.
    \item \textit{Persona-value extraction:} In the \textit{persona-value extraction} subtask, we extract persona values from the given instance for each identified persona type. These values may include a span from the input (e.g., for \textit{occupation}), another speaker present in the conversation (e.g., for \textit{relationship}), or something inferred from the context (e.g., for \textit{trait}).
\end{itemize}

We employed three annotators\footnote{They were NLP researchers or linguistics by profession; and their age ranges between $20-45$ years.} to annotate \dataset. 
The first two annotators assigned appropriate persona labels to utterances of the dialogue and any disagreement between them was resolved by the third annotator.
We measured the consistency of annotations among annotators by measuring Krippendorff's Alpha inter-annotator agreement \citep{krippendorff2011computing}.
For {persona discovery}, we obtained the 
inter-annotator
agreement score of $0.83$, while for {persona-type identification}, the agreement is $0.71$.
Table \ref{tab:data_stats} shows the dataset statistics and
presents the distribution of persona types in \dataset.


\subsection{Persona Types}
\label{sec:app_persona_type}

Persona-chat \cite{zhang2018personalizing} is a multi-turn dialogue dataset containing conversations conditioned on personas. The dataset comprises $1155$ possible artificial personas, each of which is a combination of $3$ to $5$ profile sentences such as `I like to eat cheetos,' `I am a doctor,' and `I have two large dogs.' The authors employed crowd-sourced workers to converse with each other by assuming the personality of the artificial persona provided to them. Through this process, the authors were able to collect $10907$ dialogues. We study the personas and the associated dialogues in this dataset to gain insight into the diversity of persona information. Persona-chat also includes provisional information about the speakers such as `I just got my nails done' or `I am on a diet now.' However, we disregard such persona statements as they do not contribute to the persistent persona information that we aim to uncover in our work. After careful consideration, we are able to identify five persona types that were persistent in nature. We present the persona types along with their definitions and examples in Table \ref{tab:person_slots_and_definition}.

In addition to analyzing the Persona-chat dataset, we also analyze the dialogues present in \dataset\ to determine if the defined persona-types encompass all possibilities. With the inclusion of the `miscellaneous' label, we are able to capture all conceivable persona information. Furthermore, we observe that with sufficient evidence, the `miscellaneous' label can be subdivided further into categories such as `dislikes,' `education,' `important dates,' and `important places.' However, we contend that there will always be a need for a label such as `miscellaneous' given that the possibilities for persona information are boundless.

\subsection{Annotation Guidelines}
\label{sec:app_ann_guide}
We employ three annotators to annotate the proposed dataset, \dataset.
The annotations are performed in three stages. In the first stage, the annotators are asked to identify all the utterances in each dialogue that contain any persona related information by marking them as `yes' and all others as `no'.
In the second stage, we ask the annotators to consider only the utterances marked as `yes' in the previous stage and identify the type of persona present in it (out of the five possible persona-types defined earlier). Finally, for each persona type, the annotators are asked to identify the value of the corresponding persona type. These values can either be present as a span in the input or can be a result of an inference made by the annotator. 

Each instance for the second stage is made up of a sequence of utterances $\{u_1, u_2, \cdots, u_i\}$, where the last utterance is the one marked as containing persona information in the first stage. The annotators are given the following guidelines to decide the persona-type for the utterance $u_i$:
\begin{itemize}[leftmargin=*,noitemsep,topsep=0pt]
    \item \textbf{\textit{Trait:}} Mark the utterance $u_i$ as containing trait persona information if the information gathered from utterances $\{u_1, u_2, \cdots, u_i\}$ indicates towards the speaker having any type of a distinguishing quality. 
    \item \textbf{\textit{Likes:}} The utterance $u_i$ contains persona information about likes if the utterances from $u_1$ to $u_i$ indicate the concerned speaker finds someone/something pleasant or enjoyable.
    \item \textbf{\textit{Occupation:}} If until utterance $u_i$ we have enough evidence to conclude the profession of the speaker, we mark that utterance as containing information about occupation.
    \item \textbf{\textit{Relation:}} Mark an utterance $u_i$ as having the information of the speaker's relationship if enough evidence has been found from context of them being related to someone biologically (mother, father) or otherwise (friend, spouse).
    \item \textbf{\textit{Miscellaneous:}} If the annotators find evidence of some important perpetual information related to the speaker (like education, and important dates) present in utterance $u_i$ and if this information does not fall under any other identified persona types, then they mark it as miscellaneous.
\end{itemize}

 
 
 
 


An example annotation for each of the subtasks is shown in Table \ref{tab:data_eg}. There are two speakers with four utterances in a dialogue. The first utterance does not carry any persona information; hence, it is tagged as \textit{``no"} for the {persona discovery} task, and persona-type is not applicable for the case. For the remaining three utterances, the associated persona types are \textit{occupation}, \textit{traits}, and \textit{likes}, respectively. For the corresponding persona-type, we show the persona-value in the last column of the table. While \textit{occupation} and \textit{likes} comes from the input text, the \textit{trait} comes after inferencing knowledge from the input.

%% file: tab-instigators-defn.tex
\begin{table*}[ht]
\centering
\resizebox{\textwidth}{!}{%
\begin{tabular}{l|p{22em}|p{42em}} 
\hline
\multicolumn{1}{c|}{\textbf{Persona Type}} & \multicolumn{1}{c|}{\textbf{Definition}} & \multicolumn{1}{c}{\textbf{Example}} \\ 
\hline
Trait & A quality that forms the part of the speaker's character & \em Also, I was the point person on my company's transition from the KL-5 to GR-6 system. \textcolor{blue}{(Trait: responsible)}\\ \hline
Likes & Something/someone the speaker finds pleasant & \em I hate that I have to say goodbye to Ross \textcolor{blue}{(Likes: Ross)}\\ \hline
Relationship & The way two speakers are related to each other & \em Well, if you want, I'll just-I'll just break it off with Mindy. \textcolor{blue}{(Relationship:Mindy)}\\ \hline
Occupation & Profession or Job & \em I have an audition for the lead role tomorrow! \textcolor{blue}{(Occupation: Actor)}\\ \hline
\multirow{2}{*}{Miscellaneous} & Any other persona-related information. For example, education, important days, important places & \multirow{2}{*}{\em My parents are from Lebanon. \textcolor{blue}{(Misc: Parents from Lebanon)}} \\ \hline
\end{tabular}
}
\caption{Persona types with their definitions.}
\label{tab:person_slots_and_definition}
 \vspace{-2mm}
\end{table*}

%% file: tab-dataset-instance.tex
\begin{table*}[ht]
\centering
\resizebox{\textwidth}{!}{%
\begin{tabular}{c|l|l|c|c|c}
\hline
\multirow{2}{*}{\textbf{\#}} &
\multirow{2}{*}{\textbf{Speaker}} & \multicolumn{1}{c|}{\multirow{2}{*}{\textbf{Utterance}}} & \multicolumn{3}{c}{\bf Tasks} \\ \cline{4-6}
& & & \multicolumn{1}{c|}{\textbf{Persona Discovery}} & \multicolumn{1}{c}{\textbf{Persona-type Identification}} & \multicolumn{1}{c}{\textbf{Persona-value Generation}} \\ \hline
$u_1$ & Chandler & \em What've you been up to? & No & - & - \\ \hline
$u_2$ & Jade & \em Oh, you know, the usual, teaching aerobics, partying way too much. & Yes & Occupation & Teaches aerobics \\ \hline
$u_3$ & Jade & \em Oh, and in case you were wondering, those are my legs on the new James Bond poster. & Yes & Trait & Boastful \\ \hline
$u_4$ & Chandler & \em Can you hold on a moment? I have another call.  I love her. & Yes & Likes & Jade \\ \hline
\end{tabular}%
}
\caption{Example of an annotated dialogue from \dataset.}
\label{tab:data_eg}
 \vspace{-4mm}
\end{table*}

%% file: tab-dataset-stats.tex
\begin{table}[t]
    \centering
    \resizebox{\columnwidth}{!}{%
    \begin{tabular}{|c|c|c|c|c|c|}
    \hline
    {\textbf{Set}} & {\textbf{\#Dlg}} & {\textbf{\#Utt}} & \textbf{Avg \#Sp/Dlg} & {\textbf{\#Persona Utt}} & \textbf{Avg \#Persona Utt/Dlg)} \\ \hline
    \textbf{Train} & 1039 & 9989 & \multicolumn{1}{c|}{2.70} & \multicolumn{1}{c|}{1005} & \multicolumn{1}{c|}{0.96} \\ \hline
    \textbf{Dev} & 114 & 1109 & \multicolumn{1}{c|}{3.01} & \multicolumn{1}{c|}{109} & \multicolumn{1}{c|}{0.96} \\ \hline
    \textbf{Test} & 280 & 1983 & \multicolumn{1}{c|}{2.66} & \multicolumn{1}{c|}{305} & \multicolumn{1}{c|}{1.09} \\ \hline
    \end{tabular}%
    }\\\vspace{3mm}
    \resizebox{0.7\columnwidth}{!}{%
    \begin{tabular}{|c|ccccc|}
    \hline
    \multirow{2}{*}{\textbf{Set}} & \multicolumn{5}{c|}{\textbf{\#Persona Slot}} \\ \cline{2-6} 
     & \multicolumn{1}{c|}{\textbf{Trait}} & \multicolumn{1}{c|}{\textbf{Likes}} & \multicolumn{1}{c|}{\textbf{Relation}} & \multicolumn{1}{c|}{\textbf{Occupation}} & \textbf{Misc} \\ \hline
    \textbf{Train} & \multicolumn{1}{c|}{389} & \multicolumn{1}{c|}{244} & \multicolumn{1}{c|}{107} & \multicolumn{1}{c|}{89} & 179 \\ \hline
    \textbf{Dev} & \multicolumn{1}{c|}{32} & \multicolumn{1}{c|}{36} & \multicolumn{1}{c|}{10} & \multicolumn{1}{c|}{10} & 24 \\ \hline
    \textbf{Test} & \multicolumn{1}{c|}{120} & \multicolumn{1}{c|}{88} & \multicolumn{1}{c|}{28} & \multicolumn{1}{c|}{18} & 53 \\ \hline
    \end{tabular}%
    }
    \caption{Statistics of \dataset.}
    \label{tab:data_stats}
    \vspace{-3mm}
\end{table}

%% file: 6_methodology_coling.tex
In this section, we illustrate \model, our proposed method to benchmark the task and the dataset. 
\model\ constitutes three subtasks -- the first subtask,  persona discovery, aims to identify all utterances carrying persona information from a given dialogue, the second subtask, persona-type identification, deals with identifying the exact type of persona present in the utterances, while the third task, persona-value extraction, extracts the exact persona value for each identified type.

\begin{figure*}[h!]
    \centering
    \subfloat[\textit{Persona discovery}. The utterances representation obtained from utterance-level Transformer $T_u$ are fed to the dialogue-level Transformer $T_d$ to obtain the contextual representation for each utterance which is then used for classification.]{
        \includegraphics[width=0.27\textwidth]{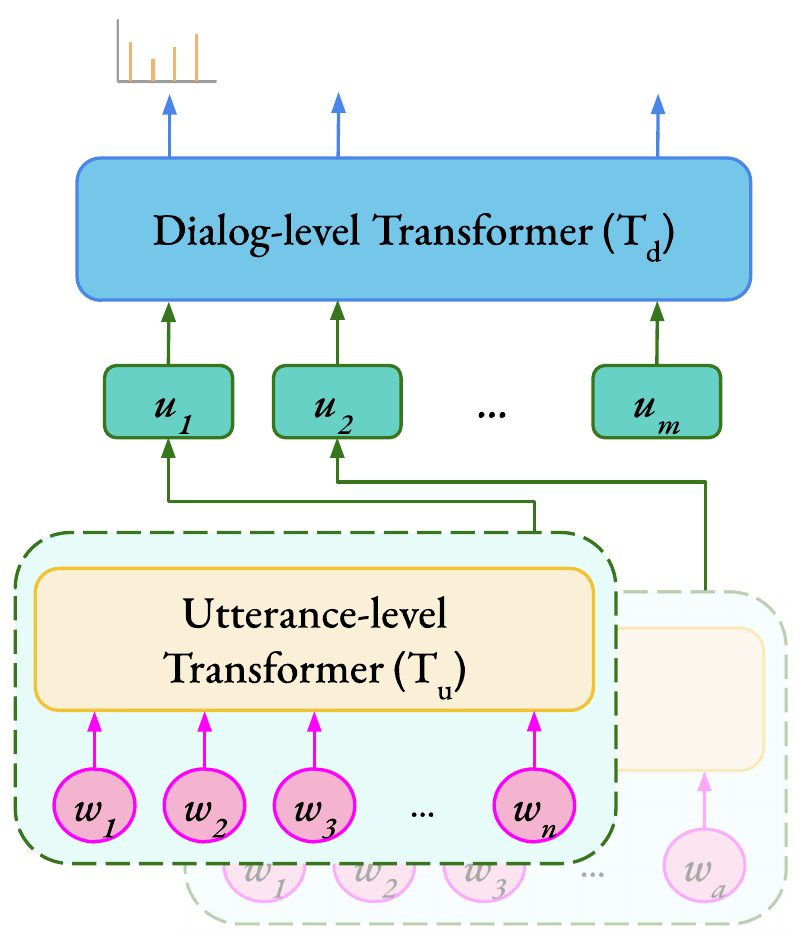}
        \label{fig:arch_model1}
     } \hfill
    \subfloat[\textit{Persona-type identification}. Utterance representation is obtained from $GRU_u$, which are then passed to the dialogue-level Transformer $T_d$ to obtain the contextual representation of each utterance. These representations are also passed to speaker-specific Transformer $T_S$. After receiving the representation from context, speaker, and global attention mechanism, the final representation is used to obtain adaptive decision boundary. We initialize the centroids $\{c_i\}^{K}_{i=1}$ and the radius of decision boundaries $\{\delta\}^{K}_{i=1}$ for each persona type and use the boundary loss for optimisation.]{
        \includegraphics[width=0.65\textwidth]{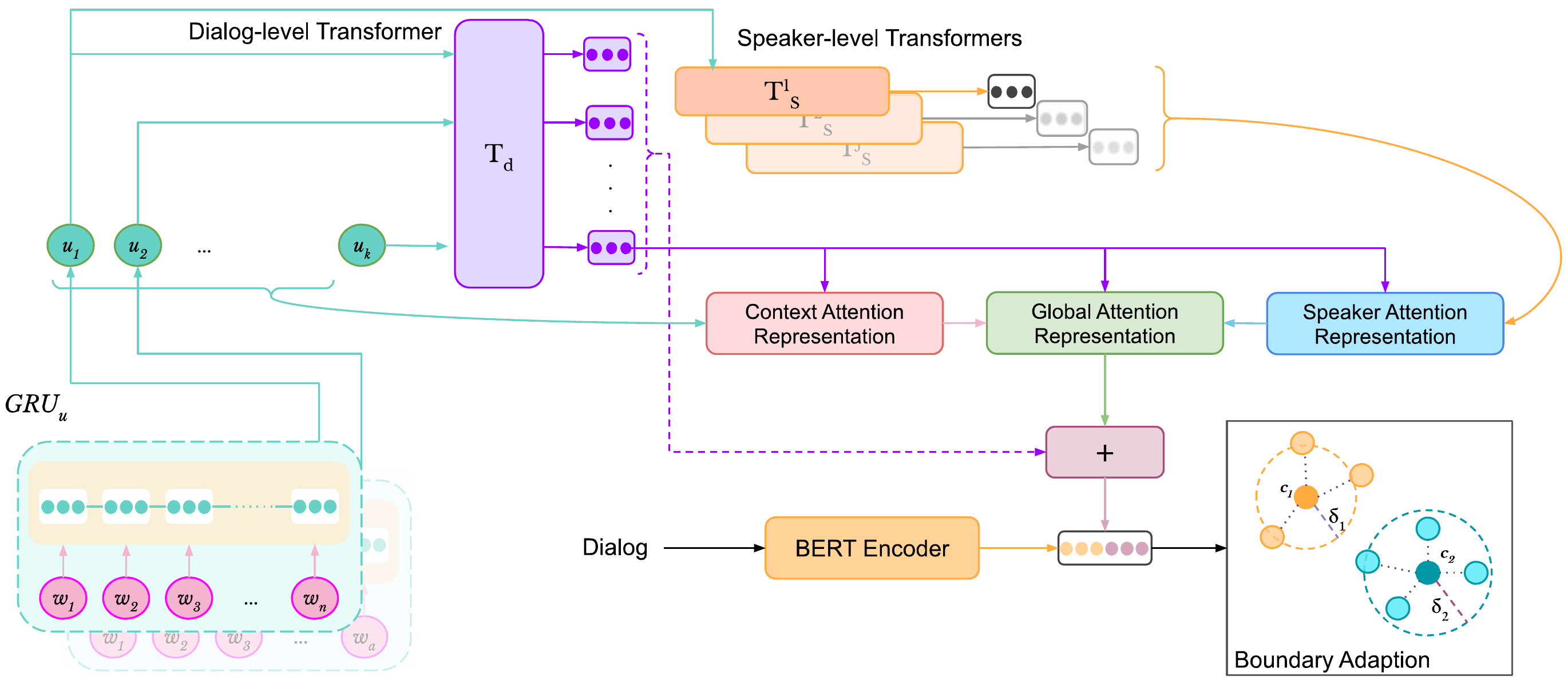}
        \label{fig:arch_model2}
    }\\
    \subfloat[\textit{Persona-value extraction}. The context, target utterance, and the complete dialogue is transformed using a BART encoder following which attention is applied to get target attended vectors. Finally a concatenated vector is sent to the BART decoder for output generation.]{
        \includegraphics[width=\textwidth]{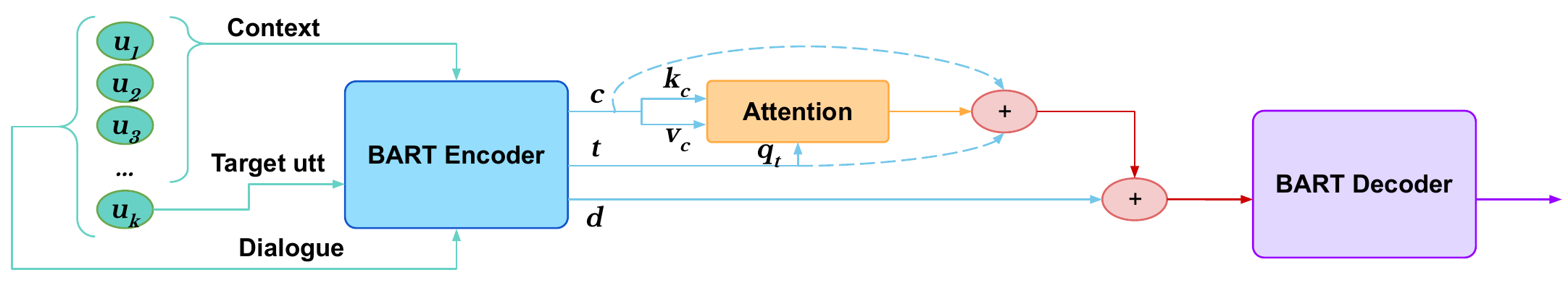}
        \label{fig:arch_model3}
    }
    \caption{Model architecture of \model.}
    \vspace{-2mm}
\end{figure*}

\subsection{Persona Discovery}
In this phase, we employ two Transformer \citep{vaswani2017attention} encoders, $T_u$ and $T_d$, to capture the word-level  and the dialogue-level contexts, respectively. Figure \ref{fig:arch_model1} shows the schematic diagram of the proposed architecture. The input to the model is a dialogue as a sequence of utterances, $D = \{u_1, u_2, ..., u_m\}$.
We pass dialogue-level representations to fully-connected layers for classification. The \dataset\ dataset is skewed towards the utterances not containing any persona information. To overcome this imbalance, we upsample the utterances carrying persona information using the SMOTE upsampling technique \citep{chawla2002smote}.

\subsection{ {Persona-type Identification}}
Now that we know which utterances of the dialogue carry persona, we pass these utterances and their context to the next subtask. This process takes as input a sequence of utterances, $I = \{u_1, u_2, ..., u_k\}$, where $u_k$ is the utterance containing persona information, i.e., the target utterance, whereas utterances, $u_1, u_2, ..., u_{k-1}$ are the contextual utterances. We propose a GRU \citep{cho2014learning} and Transformer based approach for this subtask. The proposed architecture is shown in Figure \ref{fig:arch_model2}.\\
\textbf{Dialogue representation.}
Each utterance $u_j$ of the input sequence $I$ is passed through an utterance-level BiGRU $GRU_u$ to obtain its representation $\hat{h}_u$.
This representation, $\hat{h}_u$, is passed to a dialogue-level Transformer encoder $\hat{T}_d$ to obtain a context-rich representation for each utterance $\hat{h}_d$.\\
\textbf{Speaker-specific representation.} 
To effectively capture the speaker sequence in a dialogue, we employ a separate Transformer encoder for each speaker present in the dialogue to obtain a contextual speaker-specific representation. Each speaker-specific encoder $T_S^i$ is fed with the utterance representations obtained from $\hat{T}_d$ specific to  speaker $i$.\\
\textbf{Attention.} 
We generate speaker-aware representation $H_{SAR}$ by computing attention \citep{bahdanau2014neural} between speaker-specific representations obtained from $T_S$ and the target representation obtained from $\hat{T}_d$. Similarly, we obtain the context-aware representation $H_{CAR}$.
We further compute global attention $H_{GAR}$ to fuse the speaker-aware and context-aware representations. 
Additionally, to capture the dialogue-level context effectively, we employ a pretrained BERT \citep{devlin2018bert} encoder to encode it.\\
\textbf{Adaptive decision boundary.} We learn the adaptive decision boundary for the persona classes using a $k$-means approach.
Furthermore, to ensure maximum separation of any two classes and to learn an optimal decision boundary for each class, we use the boundary loss \citep{zhang2021deep}.
The loss is computed using the following equation:
\vspace{-3mm}
\begin{equation}
\begin{split}
    L_b = \frac{1}{N}\sum_{i=1}^N[\Delta_i(||z_i - c_{yi}||_2 - \delta_{yi}) + \\
    (1-\Delta_i)(\delta_{yi} - ||z_i-c_{yi}||_2)] \nonumber
\end{split}
\end{equation}
where $N$ is the total number of samples in our set, $z_i$ is the representation of the $i^{th}$ instance, $c_{yi}$ is the centroid for class $yi$, and $\delta_{yi}$ is the radius for class $yi$. Here,\\
\vspace{-3mm}
\begin{center}
    $\Delta_i =$
    \begin{math}
      \left\{
        \begin{array}{l}
          1, if ||z_i-c_{yi}||_2 > \delta_{yi}\\
          0, if ||z_i-c_{yi}||_2 \leq \delta_{yi}
        \end{array}
      \right.
    \end{math}
\end{center}


\subsection{{Persona-value Extraction}}
Extraction of the persona values for each of the specified persona types in a dialogue is the last step in the process of creating a speaker profile. These persona values might need to be inferred from the input because they are potentially unrestricted. As a result, we model the job using an encoder-decoder architecture and a generative objective. To determine the values, we must efficiently record the entire discussion to capture the substance of the dialogue, the contextual utterances to encapsulate the contextual knowledge, and the target utterance, since it is the primary source for the persona.

A BART encoder \citep{lewis2019bart} is used to encode the context, the target, and the dialogue utterance into $c$, $t$, and $d$, respectively. We pass the context and target representations to an attention mechanism where the key $k_c$, and value $v_c$ are obtained from the context representation, and the query $q_t$ comes from the target utterance. This allows for incorporating the interaction between the target utterance with contextual utterances, and thus model the context-induced persona information in the target utterance. The final representation is then concatenated with the dialogue representation and fed to the BART decoder for output generation.

%% file: 7_results_new.tex
\subsection{Experimental Setup}
We perform experiments for all three subtasks of speaker profiling in two settings -- standalone and pipeline. Following sections present both settings.


\noindent \textbf{Standalone Evaluation.}
In this setting, the models for all phases are trained and evaluated separately. In other words, for  persona discovery, all the dialogues are passed through \model\ 
and each utterance is investigated for the presence of persona information. For  persona-type identification, we consult the ground-truth and pass only the utterances carrying persona information along with their context to the model as shown in Figure \ref{fig:arch_model2}. Finally, for persona-value extraction, we give the ground truth persona types along with the utterances carrying persona information with their context to \model\ and extract the persona-value.


\noindent \textbf{Pipeline evaluation.} In the setting, persona discovery is performed as done in standalone setup. However, for persona-type identification, we feed the model with only the utterances identified as carrying persona in the former subtask without verifying the ground-truth. Finally, the output recieved from the second subtask are given as input to the last task of persona-value extraction without grount-truth consultation.

\noindent \textbf{Baseline methods.}
\label{sec:baseline}
Since the tasks of persona discovery and persona-type identification are classification based, we adapt four classification baselines from related tasks such as emotion detection and dialogue-act identification due to their similar nature to SPC. We list the baselines with a brief description here.
\begin{itemize}[leftmargin=*]
    \item \textbf{BERT:} BERT (Bidirectional Encoder Representations from Transformers) \cite{devlin2018bert} is encoder stack of transformer architecture \cite{vaswani2017attention}. We use pre-trained BERT base and fine-tune it for our tasks.
    \item \textbf{DialogXL:} Shen et al. \cite{shen2020dialogxl} modified XLNet by changing the segment-level recurrence mechanism to an utterance-level recurrence mechanism so that XLNet could be mapped to a dialogue setting. They also incorporated dialogue-aware self-attention to capture the intra- and inter-speaker dependencies in a conversation.
    \item \textbf{Co-GAT:} Qin et al. \cite{qin2020co} 
    proposed a co-interactive graph interaction layer with cross-utterance and cross-tasks connections.
    \item \textbf{AGHMN:} Jiao et al. \cite{jiao2020real} used an attention-based GRU to monitor the flow of information through a hierarchical memory network.
    The attention weights are calculated over the contextual utterances in the conversation and combined for the final classification.
\end{itemize}

For the third subtask of persona-value extraction, we consider established sequence-to-sequence models for comparison given its generative objective.
\begin{itemize}[leftmargin=*]
    \item \textbf{RNN:} OpenNMT5 provides with an implementation of the RNN seq-to-seq architecture which we use in our study.
    \item \textbf{Transformers:} We use the standard encoder and decoder stack to generate the output \cite{vaswani2017attention}.
    \item \textbf{Pointer Generator Network (PGN):} The standard seq-to-seq architecture supporting both generation of new words as well as copying words from input \cite{see2017get}.
    \item \textbf{BART:} BART \cite{lewis2019bart} contains a bidirectional encoder and an auto-regressive decoder to create a denoising auto-encoder model.
    \item \textbf{T5:} T5 \cite{raffel2020exploring} is a seq-to-seq model trained on a mixture of unsupervised and supervised tasks.
\end{itemize}

\noindent \textbf{Evaluation metrics.} 
Since the first two subtasks are multi-class classification problems, we use F1 score as our choice of evaluation metric.
We consider F1 score of the positive class for the task of \textit{persona discovery}, while weighted F1 score is used for \textit{persona type identification}. On the other hand, since the task of \textit{persona-value extraction} follows a generative objective, we use the ROUGE \citep{lin-2004-rouge} and the BLEU \citep{papineni2002bleu} scores to gauge the performance of the systems.


\input{7_1_exp_results_new}

\subsection{Ablation Study}
\model\ captures two primary aspects of a dialogue -- the dialogue context and the speaker semantics. To capture the dialogue-level context, we use \model$_{Base}$,  containing the utterance-level GRU $GRU_u$, dialogue-level Transformer $T_d$, followed by context $H_{CAR}$, and global attention representation $H_{GAR}$. We reinforce the dialogue context by adding a BERT encoder to this architecture, \model$_{Base+BERT}$. Speaker semantics are captured by the speaker-specific Transformers $T_S$ and attention representation $H_{SAR}$. We add these modules in our final model, \model. 

To assess the importance of each module, we perform ablation over \model. The last three rows of Table \ref{tab:standalone_ph2}(b) illustrates these results. We observe that
the addition of the BERT encoder improves the performance of our model considerably ($21\%$) while the addition of speaker module improves it further ($2\%$) verifying the use of each module.

\begin{table}[t]
\centering
\resizebox{\columnwidth}{!}{
\begin{tabular}{c|ccccc|c}
\hline
\textbf{Systems} & \textbf{Trait} & \textbf{Likes} & \textbf{Relation} & \textbf{Misc} & \textbf{Occupation} & \textbf{Weighted} \\ \hline
\textbf{\model$_{Base}$} & 0.36 & 0.27 & 0.22 & 0.23 & 0.07 & 0.28 \\ 
\textbf{\model$_{Base+BERT}$} & 0.56 & 0.53 & 0.45 & 0.40 & 0.26 & 0.49 \\ \hline
\textbf{\model} & \textbf{0.59} & \textbf{0.6} & \textbf{0.46} & \textbf{0.32} & \textbf{0.28} & \textbf{0.51} \\ \hline
\end{tabular}%
}
\caption{Ablation results for \textit{persona-type identification}.}
\label{tab:ablation}
\vspace{-5mm}
\end{table}

%% file: 7_1_exp_results_new.tex
\subsection{Results}

\noindent \textbf{Standalone evaluation.}
We evaluate \model\ for all subtasks of SPC separately and show the results in Table \ref{tab:standalone_ph2} and Table \ref{tab:results-phase3}.
\begin{itemize}[leftmargin=*,noitemsep,topsep=0pt]
    \item \textit{Persona discovery}: We train \model\ as a binary classifier using cross-entropy loss. We obtain $38\%$ F1 score, which is $\sim7\%$ better than the best baseline, DialogXL, as can be seen in Table \ref{tab:standalone_ph2}. The gain in performance can be attributed to the efficient way we use different modules in our architecture to capture different essence of a conversation. It can also be seen that \model\ outperforms the worst performing baseline, AGHMN, by $\sim17\%$. While our model performs the best in terms of F1 score, DialogXL and Co-GAT produce the best performance in terms of precision and recall scores, respectively. On the other hand, \model\ holds a balance between precision and recall to obtain the highest F1 score, which is our metric of choice due to the class imbalance present in our data.
    \item \textit{Persona-type identification}: We use boundary loss \cite{zhang2021deep} to train \model\ for this task. Table \ref{tab:standalone_ph2} shows that \model\ yields a weighted average of $51\%$ F1-score with the maximum score for persona-type \textit{likes}. AGHMN, the best baseline, results in a weighted average of $48\%$ F1-score, which is $\sim3\%$ less than \model. We observe that \model\ achieves the best F1-score for all persona slots showcasing a global dominance of our system. We observe a minimum improvement of $\sim3\%$, $\sim2\%$, $\sim8\%$, $\sim6\%$, and $\sim3\%$ F1-score for the persona slot -- trait, likes, relation, misc, and occupation, respectively, over the existing baselines. It is interesting to observe that our model performs quite well across the persona slots that are dominantly present in our data and consistently decreases for the slots based on their availability in the data.
    \item \textit{Persona-value extraction}: Using a generative objective, we obtain the results by \model\ for this subtask. Table \ref{tab:results-phase3} demonstrates that \model\ outperforms the baselines by around $1\%$ for all metrics except BLEU 1 and BLEU 3.
\end{itemize}

\begin{table}[t]
\centering
\resizebox{\columnwidth}{!}{%
\begin{tabular}{|c||ccc||cccccc|}
\hline
\multirow{2}{*}{\textbf{Systems}} & \multicolumn{3}{c||}{\textbf{\textit{Persona Discovery}}} & \multicolumn{6}{c|}{\textbf{\textit{Persona-type Identification}}} \\ \cline{2-10} 
 & \multicolumn{1}{c|}{\textbf{P}} & \multicolumn{1}{c|}{\textbf{R}} & \textbf{F1} & \multicolumn{1}{c|}{\textbf{Trait}} & \multicolumn{1}{c|}{\textbf{Likes}} & \multicolumn{1}{c|}{\textbf{Rel}} & \multicolumn{1}{c|}{\textbf{Misc}} & \multicolumn{1}{c|}{\textbf{Occ}} & \textbf{Wtd} \\ \hline \hline
\textbf{BERT} & \multicolumn{1}{c|}{0.17} & \multicolumn{1}{c|}{0.72} & 0.27 & \multicolumn{1}{c|}{0.48} & \multicolumn{1}{c|}{0.0} & \multicolumn{1}{c|}{0.09} & \multicolumn{1}{c|}{0.24} & \multicolumn{1}{c|}{0.05} & 0.24 \\ \hline
\textbf{DialogXL} & \multicolumn{1}{c|}{\textbf{0.45}} & \multicolumn{1}{c|}{0.23} & 0.31 & \multicolumn{1}{c|}{0.52} & \multicolumn{1}{c|}{0.0} & \multicolumn{1}{c|}{0.0} & \multicolumn{1}{c|}{0.0} & \multicolumn{1}{c|}{0.0} & 0.18 \\ \hline
\textbf{Co-GAT} & \multicolumn{1}{c|}{0.15} & \multicolumn{1}{c|}{\textbf{0.94}} & 0.27 & \multicolumn{1}{c|}{0.50} & \multicolumn{1}{c|}{0.35} & \multicolumn{1}{c|}{0.06} & \multicolumn{1}{c|}{0.14} & \multicolumn{1}{c|}{0.05} & 0.33 \\ \hline
\textbf{AGHMN} & \multicolumn{1}{c|}{0.43} & \multicolumn{1}{c|}{0.14} & 0.21 & \multicolumn{1}{c|}{0.56} & \multicolumn{1}{c|}{0.58} & \multicolumn{1}{c|}{0.38} & \multicolumn{1}{c|}{0.26} & \multicolumn{1}{c|}{0.25} & 0.48 \\ \hline \hline
\textbf{\model} & \multicolumn{1}{c|}{0.30} & \multicolumn{1}{c|}{0.50} & \textbf{0.38} & \multicolumn{1}{c|}{\textbf{0.59}} & \multicolumn{1}{c|}{\textbf{0.60}} & \multicolumn{1}{c|}{\textbf{0.46}} & \multicolumn{1}{c|}{\textbf{0.32}} & \multicolumn{1}{c|}{\textbf{0.28}} & \textbf{0.51} \\ \hline
\end{tabular}%
}
\caption{Comparative results for standalone evaluation. (P: Precison; R: Recall; Rel: Relationship; Occ: Occupation; Wtd: Weighted F1 score.}
\label{tab:standalone_ph2}
\vspace{-4mm}
\end{table}


\begin{table}[t]
\centering
\resizebox{\columnwidth}{!}{%
\begin{tabular}{|c||ccc||cccccc|}
\hline
\multirow{2}{*}{\textbf{Systems}} & \multicolumn{3}{c||}{\textbf{\textit{Persona Discovery}}} & \multicolumn{6}{c|}{\textbf{\textit{Persona-slot Identification}}} \\ \cline{2-10}
 & \multicolumn{1}{c|}{\textbf{P}} & \multicolumn{1}{c|}{\textbf{R}} & \textbf{F1} & \multicolumn{1}{c|}{\textbf{Trait}} & \multicolumn{1}{c|}{\textbf{Likes}} & \multicolumn{1}{c|}{\textbf{Rel}} & \multicolumn{1}{c|}{\textbf{Misc}} & \multicolumn{1}{c|}{\textbf{Occ}} & \textbf{Wtd} \\ \hline \hline
\textbf{BERT} & \multicolumn{1}{c|}{0.17} & \multicolumn{1}{c|}{0.72} & 0.27 & \multicolumn{1}{c|}{0.12} & \multicolumn{1}{c|}{0.0} & \multicolumn{1}{c|}{0.05} & \multicolumn{1}{c|}{0.07} & \multicolumn{1}{c|}{0.0} & 0.07 \\ \hline
\textbf{DialogXL} & \multicolumn{1}{c|}{\textbf{0.45}} & \multicolumn{1}{c|}{0.23} & 0.31 & \multicolumn{1}{c|}{0.33} & \multicolumn{1}{c|}{0.0} & \multicolumn{1}{c|}{0.0} & \multicolumn{1}{c|}{0.0} & \multicolumn{1}{c|}{0.0} & 0.07 \\ \hline
\textbf{Co-GAT} & \multicolumn{1}{c|}{0.15} & \multicolumn{1}{c|}{\textbf{0.94}} & 0.27 & \multicolumn{1}{c|}{0.31} & \multicolumn{1}{c|}{0.26} & \multicolumn{1}{c|}{0.0} & \multicolumn{1}{c|}{0.07} & \multicolumn{1}{c|}{0.0} & 0.15 \\ \hline
\textbf{AGHMN} & \multicolumn{1}{c|}{0.43} & \multicolumn{1}{c|}{0.14} & 0.21 & \multicolumn{1}{c|}{0.48} & \multicolumn{1}{c|}{0.43} & \multicolumn{1}{c|}{\textbf{0.41}} & \multicolumn{1}{c|}{0.21} & \multicolumn{1}{c|}{0.16} & 0.40 \\ \hline \hline
\textbf{\model} & \multicolumn{1}{c|}{0.30} & \multicolumn{1}{c|}{0.50} & \textbf{0.38} & \multicolumn{1}{c|}{\textbf{0.56}} & \multicolumn{1}{c|}{\textbf{0.50}} & \multicolumn{1}{c|}{0.35} & \multicolumn{1}{c|}{\textbf{0.26}} & \multicolumn{1}{c|}{\textbf{0.21}} & \textbf{0.43} \\ \hline
\end{tabular}%
}
\caption{Comparative results for pipeline evaluation.}
\label{tab-e2e}
\vspace{-4mm}
\end{table}

\noindent \textbf{Pipeline evaluation.} Tables \ref{tab-e2e} and  \ref{tab:results-phase3} show the performance obtained by our model along with the baseline scores. For the task of persona discovery, we obtain the same results as standalone due to the same type of input and evaluation strategies. However, we observe a performance drop of $\sim8\%$ for the persona-type identification task and a drop of at most $\sim6\%$ for the persona-value extraction task when compared to the standalone results. This is expected as the erroneous predictions from the previous stage may propagate to the next stage. Nevertheless, when compared to the baseline systems, our proposed mechanism gives the best score, with an increase of $\sim3\%$ over the best baseline and $\sim36\%$ over the worst performing baseline, BERT, for the former two tasks (c.f. Table \ref{tab-e2e}). Apart from \textit{relation}, \model\ performs the best for all persona slots with a performance increase of $\sim8\%$, $\sim7\%$, $\sim5\%$, and $\sim5\%$ for the persona slot -- trait, likes, misc, and occupation, respectively. While for the last subtaks, we obtain an improvement of $\sim1\%$ over the baselines. Consequently, we establish that \model\ is able to capture the essence of persona more clearly when compared with the baseline systems.

\begin{table}[t]
\centering
\resizebox{\columnwidth}{!}{%
\begin{tabular}{|c||ccccc||ccccc|}
\hline
\multirow{2}{*}{\textbf{Models}} & \multicolumn{5}{c||}{\textbf{Standalone}} & \multicolumn{5}{c|}{\textbf{Pipeline}} \\ \cline{2-11} 
 & \multicolumn{1}{c|}{\textbf{R1}} & \multicolumn{1}{c|}{\textbf{R2}} & \multicolumn{1}{c|}{\textbf{B1}} & \multicolumn{1}{c|}{\textbf{B2}} & \textbf{B3} & \multicolumn{1}{c|}{\textbf{R1}} & \multicolumn{1}{c|}{\textbf{R2}} & \multicolumn{1}{c|}{\textbf{B1}} & \multicolumn{1}{c|}{\textbf{B2}} & \textbf{B3} \\ \hline \hline
\textbf{RNN} & \multicolumn{1}{c|}{26.85} & \multicolumn{1}{c|}{2.28} & \multicolumn{1}{c|}{24.78} & \multicolumn{1}{c|}{1.48} & 0.36 & \multicolumn{1}{c|}{19.64} & \multicolumn{1}{c|}{0.37} & \multicolumn{1}{c|}{18.98} & \multicolumn{1}{c|}{0.36} & 1.12 \\ \hline
\textbf{Transformer} & \multicolumn{1}{c|}{26.02} & \multicolumn{1}{c|}{2.37} & \multicolumn{1}{c|}{23.93} & \multicolumn{1}{c|}{1.54} & 0.58 & \multicolumn{1}{c|}{19.48} & \multicolumn{1}{c|}{0.70} & \multicolumn{1}{c|}{18.82} & \multicolumn{1}{c|}{0.69} & 2.05 \\ \hline
\textbf{PGN} & \multicolumn{1}{c|}{24.40} & \multicolumn{1}{c|}{1.59} & \multicolumn{1}{c|}{23.12} & \multicolumn{1}{c|}{1.08} & 0.36 & \multicolumn{1}{c|}{17.11} & \multicolumn{1}{c|}{0.49} & \multicolumn{1}{c|}{16.13} & \multicolumn{1}{c|}{0.33} & \textbf{9.28} \\ \hline
\textbf{BART} & \multicolumn{1}{c|}{28.93} & \multicolumn{1}{c|}{2.16} & \multicolumn{1}{c|}{\textbf{27.23}} & \multicolumn{1}{c|}{1.51} & 0.36 & \multicolumn{1}{c|}{22.41} & \multicolumn{1}{c|}{\textbf{1.01}} & \multicolumn{1}{c|}{21.37} & \multicolumn{1}{c|}{0.80} & 0.08 \\ \hline
\textbf{T5} & \multicolumn{1}{c|}{15.07} & \multicolumn{1}{c|}{0.0} & \multicolumn{1}{c|}{14.90} & \multicolumn{1}{c|}{2.25} & {\textbf{1.20}} & \multicolumn{1}{c|}{11.62} & \multicolumn{1}{c|}{0.37} & \multicolumn{1}{c|}{11.51} & \multicolumn{1}{c|}{0.36} & 1.12 \\ \hline \hline
\textbf{\model} & \multicolumn{1}{c|}{\textbf{29.51}} & \multicolumn{1}{c|}{\textbf{2.97}} & \multicolumn{1}{c|}{27.16} & \multicolumn{1}{c|}{\textbf{2.27}} & 0.60 & \multicolumn{1}{c|}{\textbf{23.40}} & \multicolumn{1}{c|}{0.60} & \multicolumn{1}{c|}{\textbf{22.12}} & \multicolumn{1}{c|}{\textbf{1.12}} & 0.08 \\ \hline
\end{tabular}%
}
\caption{Comparative results for standalone and pipeline evaluation for \textit{persona-value extraction}.}
\label{tab:results-phase3}
\vspace{-4mm}
\end{table}

%% file: 8_result_analysis_new.tex
In this section, we present a detailed analysis of the results obtained for \model. We first show the quantitative analysis by analysing the confusion matrices obtained. After this, we show a qualitative analysis by observing a few test samples and their predicted persona slots. We also pick some predicted examples to illustrate the shortcomings of our approach and give a direction for future research.

\input{error_new}

\input{tab-confusion-matrix}

\subsection{Quantitative Analysis}
\noindent \textbf{\textit{{Persona discovery.}}} Table \ref{tab:confusion_mat_ph1} presents the confusion matrices for \model\ and the best performing baseline, DialogXL.
\model\ correctly predicts $184$ out of $305$ positive instances ($60.3\%$) while DialogXL is only able to predict $70$ ($22.9\%$).
Although DialogXL performs poorly while identifying the true positives, it does a better job in identifying the true negatives. It is able to correctly classify $1594$ instances as negative ($94.9\%$), whereas \model\ predicts only $1191$ true negative instances ($70.9\%$).

\noindent \textbf{\textit{Persona-type identification.}} We compare the confusion matrices obtained by \model\ and the best baseline for this subtask, AGHMN in Table \ref{tab:confusion_mat_ph2}. Both the models produce a comparable performance with maximum accurate predictions for the persona-type \textit{trait}. Moreover, we observe that the classes \textit{likes} and \textit{trait} are most confused, followed by \textit{likes}, \textit{relation}, \textit{misc} and \textit{occupation}
for \model\ as well as for AGHMN, while \textit{occupation} and \textit{relation} are least confused.

\input{tab-confusion-matrix-2}

\subsection{Qualitative Analysis}
This section presents a subjective analysis of the quality of predictions made by \model\ and the best baselines, based on a sample dialogue from the test set. The dialogue contains six utterances, where utterances $u_1$, $u_3$, and $u_5$ are identified as having the persona type \textit{relationship}, as shown in Table \ref{tab:error}. In the first subtask of persona discovery, \model\ correctly identifies two positive instances out of the total three, while the best baseline, DialogXL, only identifies one such instance. However, both \model\ and DialogXL misclassify one utterance as false negative.

Moving on to the second subtask of persona-type identification, \model\ correctly classifies two instances of persona-type \textit{relationship} but misclassifies one instance as \textit{trait}. On the other hand, the best baseline, AGHMN, only predicts one correct class and misclassifies the others as \textit{likes}.

\subsection{Common Errors by {\model}}
\input{false_positives_new}
\input{false_negatives_new}

\textbf{\textit{{False positives.}}}
While attaining a decent value for true positives,
\model\ obtains a significant value of false positives ($487$) for \textit{persona discovery} (c.f. Table \ref{tab:confusion_mat_ph1}). We analyse the type of misclassified instances and observe that \model\ often identifies utterances containing questions as having persona information.
For example, the utterance {\em `We're in a relationship?'} is marked as a positive instance for persona discovery
when in true sense, its answer was the one carrying persona.
Table \ref{tab:false_positives_egs} presents similar examples.
This phenomenon can be attributed to the presence of words such as `relationship' (instance $1$), `like' (instance $2$), or `father' (instance $4$) in the utterances as these words may hint towards the presence of explicit persona information in a statement.
In addition, \model\ frequently predicts the utterances expressing  a temporary/trivial state for the speaker as containing persona information.
For instance, the utterance {\em `Oh my god, I am losing my mind.'} is marked as the one containing persona information.
We show more such instances in Table \ref{tab:false_positives_egs}.
Future work could be done to handle such cases of false positives.

\textbf{\textit{{False negatives.}}}
In addition to falsely identifying utterances containing no persona information as positive instances, 
\model\ identifies $121$ true positive instances as belonging to the negative class. We analyse the misclassified positive instances and identify two situations where such misclassifications happen. When the persona information is present in the answer to a question, it is often misclassified by \model.
For example, in the dialogue {\em `\textit{Ross}: Okay! All right! Now, Chandler you-you wanna live with Monica, right? \textit{Chandler}: Yeah, I do.'}, Chandler's utterance contains information about his persona (\textit{relationship} with Monica), but \model\ is unable to identify this instance correctly. Table \ref{tab:false_negatives_egs} highlights similar examples from \dataset.
Furthermore, \model\ often misclassifies instances where the persona information is implicit in nature.
For instance, the utterance {\em `Ya see, it’s just, see I was a regular on a soap opera y’know?'} contains persona information (that the speaker's \textit{occupation} is an actor); however, \model\ is not able to relate the phrase `soap opera' to \textit{occupation} and thus does not mark the instance as having persona information.
Supporting examples are shown in Table \ref{tab:false_negatives_egs}.


%% file: error_new.tex
\begin{table*}[t]
\centering
\resizebox{\textwidth}{!}{
{%
\begin{tabular}{|l|l|p{20em}|c|c|c|c|c|c|}
\hline
\multicolumn{1}{|c|}{\multirow{3}{*}{\textbf{\#}}} & \multicolumn{1}{c|}{\multirow{3}{*}{\textbf{Speaker}}} & \multicolumn{1}{c|}{\multirow{3}{*}{\textbf{Utterance}}} & \multicolumn{3}{c|}{\textbf{\textit{Persona Discovery}}} & \multicolumn{3}{c|}{\textbf{\textit{Persona Type Identification}}} \\ \cline{4-9} 
\multicolumn{1}{|c|}{} & \multicolumn{1}{c|}{} & \multicolumn{1}{c|}{} & \multicolumn{1}{c|}{\multirow{2}{*}{\textbf{True}}} & \multicolumn{2}{c|}{\textbf{Predicted}} & \multicolumn{1}{c|}{\multirow{2}{*}{\textbf{True}}} & \multicolumn{2}{c|}{\textbf{Predicted}} \\ \cline{5-6} \cline{8-9} 
\multicolumn{1}{|c|}{} & \multicolumn{1}{c|}{} & \multicolumn{1}{c|}{} & \multicolumn{1}{c|}{} & \multicolumn{1}{c|}{\textbf{\model}} & \multicolumn{1}{c|}{\textbf{DialogXL}} & \multicolumn{1}{c|}{} & \multicolumn{1}{c|}{\textbf{\model}} & \multicolumn{1}{c|}{\textbf{AGHMN}} \\ \hline
\rowcolor{LightCyan} $u_1$ & Rachel & Everybody, this is Paolo, Paolo, I want you to meet my friends. This is Monica & Yes & Yes & \textcolor{blue}{No} & relationship & relationship & \textcolor{blue}{likes} \\ \hline
$u_2$ & Monica & Hi! & No & No & No & - & - &  \\ \hline
\rowcolor{LightCyan} $u_3$ & Rachel & And Joey... & Yes & \textcolor{blue}{No} & Yes & relationship & relationship & relationship \\ \hline
$u_4$ & Monica & Hi! & No & No & No & - & - &  \\ \hline
\rowcolor{LightCyan} $u_5$ & Rachel & And Ross... & Yes & Yes & Yes & relationship & \textcolor{blue}{trait} & \textcolor{blue}{likes} \\ \hline
$u_6$ & Monica & Hi! & No & No & \textcolor{blue}{Yes} & - & - &  \\ \hline
\end{tabular}%
}}
\caption{Actual and predicted labels for the \textit{persona discovery} and \textit{persona type identification} tasks. DialogXL and AGHM are the best performing baseline for the respective tasks.}
\label{tab:error}
\vspace{-4mm}
\end{table*}

%% file: tab-confusion-matrix.tex
\begin{table}[t]
\centering
\resizebox{0.7\columnwidth}{!}
{%
\subfloat[\model]{
\begin{tabular}{cccc}
 &  & \multicolumn{2}{c}{\textbf{Predicted}} \\ \cline{3-4} 
 & \multicolumn{1}{c|}{} & \multicolumn{1}{c|}{\textit{No}} & \multicolumn{1}{c|}{\textit{Yes}} \\ \cline{2-4} 
\multicolumn{1}{c|}{\multirow{2}{*}{\rotatebox{90}{\textbf{True}}}} & \multicolumn{1}{c|}{\textit{No}} & \multicolumn{1}{c|}{1191} & \multicolumn{1}{c|}{487} \\ \cline{2-4} 
\multicolumn{1}{c|}{} & \multicolumn{1}{c|}{\textit{Yes}} & \multicolumn{1}{c|}{121} & \multicolumn{1}{c|}{\textbf{184}} \\ \cline{2-4} 
\end{tabular}}
\subfloat[DialogXL]{
\begin{tabular}{cccc}
 &  & \multicolumn{2}{c}{\textbf{Predicted}} \\ \cline{3-4} 
 & \multicolumn{1}{c|}{} & \multicolumn{1}{c|}{\textit{No}} & \multicolumn{1}{c|}{\textit{Yes}} \\ \cline{2-4} 
\multicolumn{1}{c|}{\multirow{2}{*}{\rotatebox{90}{\textbf{True}}}} & \multicolumn{1}{c|}{\textit{No}} & \multicolumn{1}{c|}{1594} & \multicolumn{1}{c|}{84} \\ \cline{2-4} 
\multicolumn{1}{c|}{} & \multicolumn{1}{c|}{\textit{Yes}} & \multicolumn{1}{c|}{235} & \multicolumn{1}{c|}{\textbf{70}} \\ \cline{2-4} 
\end{tabular}}%
}
\caption{Confusion matrices for \model\ and DialogXL (best baseline) for the persona discovery task.}
\label{tab:confusion_mat_ph1}
\vspace{-5mm}
\end{table}

%% file: tab-confusion-matrix-2.tex
\begin{table}[t!]
\centering
\resizebox{\columnwidth}{!}{%
\begin{tabular}{cl|ccccc|}
\multicolumn{2}{l}{\multirow{2}{*}{}} & \multicolumn{5}{c}{\textbf{Predicted}} \\ 
\cline{3-7}
\multicolumn{2}{l|}{} & \multicolumn{1}{l|}{Trait} & \multicolumn{1}{l|}{Occupation} & \multicolumn{1}{l|}{Misc} & \multicolumn{1}{l|}{Likes} & Relation \\ 
\cline{2-7}
\multicolumn{1}{c|}{\multirow{5}{*}{\rotatebox{90}{\textbf{True}}}} & Trait & \multicolumn{1}{c|}{70/{\color{red}65}} & \multicolumn{1}{c|}{9/11} & \multicolumn{1}{c|}{10/10} & \multicolumn{1}{c|}{23/23} & 6/9 \\ 
\cline{2-2}
\cdashline{3-7}
\multicolumn{1}{c|}{} & Occupation & \multicolumn{1}{c|}{5/5} & \multicolumn{1}{c|}{6/6} & \multicolumn{1}{c|}{3/3} & \multicolumn{1}{c|}{3/3} & 1/1 \\ 
\cline{2-2}
\cdashline{3-7}
\multicolumn{1}{c|}{} & Misc & \multicolumn{1}{c|}{17/16} & \multicolumn{1}{c|}{7/8} & \multicolumn{1}{c|}{15/{\color{red}12}} & \multicolumn{1}{c|}{5/7} & 9/10 \\ 
\cline{2-2}
\cdashline{3-7}
\multicolumn{1}{c|}{} & Likes & \multicolumn{1}{c|}{24/25} & \multicolumn{1}{c|}{2/5} & \multicolumn{1}{c|}{4/4} & \multicolumn{1}{c|}{53/{\color{red}51}} & 5/3 \\ 
\cline{2-2}
\cdashline{3-7}
\multicolumn{1}{c|}{} & Relation & \multicolumn{1}{c|}{3/3} & \multicolumn{1}{c|}{0/0} & \multicolumn{1}{c|}{7/8} & \multicolumn{1}{c|}{3/5} & 15/{\color{red}12} \\ 
\cline{2-7}
\end{tabular}%
}
\caption{Confusion matrices for the \textit{persona-type identification} task. Each cell represents value like \{\model/AGHM\}.}
\label{tab:confusion_mat_ph2}
\vspace{-5mm}
\end{table}

%% file: false_positives_new.tex
\begin{table}[h!]
\centering
\subfloat[Persona does not lie in questions.]{
\resizebox{\columnwidth}{!}{%
\begin{tabular}{|l|l|p{20em}|c|c|}
\hline
\multicolumn{1}{|c|}{\multirow{2}{*}{\textbf{\#}}} & \multicolumn{1}{c|}{\multirow{2}{*}{\textbf{Speaker}}} & \multicolumn{1}{c|}{\multirow{2}{*}{\textbf{Utterance}}} & \multicolumn{2}{c|}{\textbf{Persona discovery}} \\ \cline{4-5} 
\multicolumn{1}{|c|}{} & \multicolumn{1}{c|}{} & \multicolumn{1}{c|}{} & \textbf{True} & \textbf{Predicted} \\ \hline
1 & Chandler & We're in a relationship? & 0 & {\color{red}1} \\ \hline
2 & Danny & So you like the short hair better? & 0 & {\color{red}1} \\ \hline
3 & Rachel & Yeah. Oh! Was how you invented the cotton gin?! & 0 & {\color{red}1} \\ \hline
4 & Phoebe & Well, so, umm, anyway umm, I’ve been, I’ve been looking for my Father, and umm, have you heard from him, or seen him? & 0 & {\color{red}1} \\ \hline
5 & Janice & So, I hear, you hate me? & 0 & {\color{red}1} \\ \hline
\end{tabular}%
}
}

\subfloat[Persona is not temporary/trivial attributes.]{
\resizebox{\columnwidth}{!}{%
\begin{tabular}{|l|l|p{20em}|c|c|}
\hline
\multicolumn{1}{|c|}{\multirow{2}{*}{\textbf{\#}}} & \multicolumn{1}{c|}{\multirow{2}{*}{\textbf{Speaker}}} & \multicolumn{1}{c|}{\multirow{2}{*}{\textbf{Utterance}}} & \multicolumn{2}{c|}{\textbf{Persona discovery}} \\ \cline{4-5} 
\multicolumn{1}{|c|}{} & \multicolumn{1}{c|}{} & \multicolumn{1}{c|}{} & \textbf{True} & \textbf{Predicted} \\ \hline
1 & Monica & Oh my god, I am losing my mind. & 0 & {\color{red}1} \\ \hline
2 & Phoebe & Because we’re girls. & 0 & {\color{red}1} \\ \hline
3 & Monica & No, Phoebe, I’ll tell you what, if you get ready now I'll let you play it at the wedding. & 0 & {\color{red}1} \\ \hline
4 & Leslie & My best shoes, so good to me. & 0 & {\color{red}1} \\ \hline
5 & Chandler & Okay uh, for now, temporarily, you can call me, Clint. & 0 & {\color{red}1} \\ \hline
\end{tabular}%
}}

\caption{Examples of false positives by \model.}
\label{tab:false_positives_egs}
\vspace{-2mm}
\end{table}

%% file: false_negatives_new.tex
\begin{table}[ht!]
\centering
\subfloat[Persona lies in answer to a question.\label{tab:false_negative_answers}]{
\resizebox{\columnwidth}{!}{%
\begin{tabular}{|l|l|l|p{20em}|c|c|}
\hline
\multicolumn{1}{|c|}{\multirow{2}{*}{\textbf{\#}}} & \multicolumn{1}{c|}{\multirow{2}{*}{\textbf{\#}}} & \multicolumn{1}{c|}{\multirow{2}{*}{\textbf{Speaker}}} & \multicolumn{1}{c|}{\multirow{2}{*}{\textbf{Utterance}}} & \multicolumn{2}{c|}{\textbf{Persona discovery}} \\ \cline{5-6} 
\multicolumn{1}{|c|}{} & \multicolumn{1}{c|}{} & \multicolumn{1}{c|}{} & \multicolumn{1}{c|}{} & \multicolumn{1}{c|}{\textbf{True}} & \multicolumn{1}{c|}{\textbf{Predicted}} \\ \hline
\multirow{2}{*}{1} & $u_1$ & Ross & Okay! All right! Now, Chandler you-you wanna live with Monica, right? & 0 & {\color{red}1} \\ \cline{2-6} 
 & \cellcolor{LightCyan} $u_2$ & \cellcolor{LightCyan} Chandler & \cellcolor{LightCyan} Yeah, I  do. & \cellcolor{LightCyan} 1 & \cellcolor{LightCyan} {\color{red}0} \\ \hline
\multirow{2}{*}{2} & $u_1$ & Judge & So based on your petition you are seeking an annulment on the grounds that Mr. Geller is mentally unstable? & 1 & 1 \\ \cline{2-6} 
 & \cellcolor{LightCyan} $u_2$ & \cellcolor{LightCyan} Ross & \cellcolor{LightCyan} Fine, I’m mentally unstable. & \cellcolor{LightCyan} 1 & \cellcolor{LightCyan} {\color{red}0} \\ \hline
\multirow{2}{*}{3} & $u_1$ & Ross & Are you intrigued? & 0 & 0 \\ \cline{2-6} 
 & \cellcolor{LightCyan} $u_2$ & \cellcolor{LightCyan} Chandler & \cellcolor{LightCyan} You're flingin'-flangin' right I am! & \cellcolor{LightCyan} 1 & \cellcolor{LightCyan} {\color{red}0} \\ \hline
\multirow{2}{*}{4} & $u_1$ & Rachel & Why, does she have a bad personality? & 1 & 1 \\ \cline{2-6} 
 & \cellcolor{LightCyan} $u_2$ & \cellcolor{LightCyan} Phoebe & \cellcolor{LightCyan} Oh no, Bonnie’s the best! & \cellcolor{LightCyan} 1 & \cellcolor{LightCyan} {\color{red}0} \\ \hline
\multirow{2}{*}{5} & $u_1$ & Chandler & Soo, ah, Eric, what kind of photography do ya do? & 0 & 0 \\ \cline{2-6} 
 & \cellcolor{LightCyan} $u_2$ & \cellcolor{LightCyan} Eric & \cellcolor{LightCyan} Oh, mostly fashion, so there may be models here from time to time, I hope that’s cool. & \cellcolor{LightCyan} 1 & \cellcolor{LightCyan} {\color{red}0} \\ \hline
\end{tabular}%
}}

\subfloat[Persona is implicit.\label{tab:false_negative_implicit}]{
\resizebox{\columnwidth}{!}{%
\begin{tabular}{|l|l|p{20em}|c|c|}
\hline
\multicolumn{1}{|c|}{\multirow{2}{*}{\textbf{\#}}} & \multicolumn{1}{c|}{\multirow{2}{*}{\textbf{Speaker}}} & \multicolumn{1}{c|}{\multirow{2}{*}{\textbf{Utterance}}} & \multicolumn{2}{c|}{\textbf{Persona discovery}} \\ \cline{4-5} 
\multicolumn{1}{|c|}{} & \multicolumn{1}{c|}{} & \multicolumn{1}{c|}{} & \textbf{True} & \textbf{Predicted} \\ \hline
1 & Joey & Ya see, it’s just, see I was a regular on a soap opera y’know? & 1 & {\color{red}0} \\ \hline
2 & Joey & Awww, one of my students got an audition. I’m so proud. & 1 & {\color{red}0} \\ \hline
3 & Joey & Yeah but we won’t be able to like get up in the middle of the night and have those long talks about our feelings and the future. & 1 & {\color{red}0} \\ \hline
4 & Janice & Oh, Chandler, look. You and Monica are meant to have children.    I am sure it's gonna be just fine. & 1 & {\color{red}0} \\ \hline
5 & Steve & Umm, see, I was thinking maybe you two could switch apartments because Phoebe's more our kind of people. & 1 & {\color{red}0} \\ \hline
\end{tabular}}%
}
\caption{Examples of false negatives by \model.}
\label{tab:false_negatives_egs}
\vspace{-3mm}
\end{table}

%% file: 9_conclusion_new.tex
In this paper, we explored the task of speaker profiling in conversations, aimed to discover persona information associated with each speaker involved in a conversation. We divided the task into three subtasks: persona discovery, persona-type identification, and persona-value extraction. 
We curated a new dataset, \dataset\ to benchmark this task. Furthermore, we proposed \model, a deep neural approach for the three subtasks that used Transformer and GRU modules to capture the dialogue context and speaker semantics effectively. In addition, we adapted several state-of-the-art models for comparison. Through extensive experiments, we showed the dominance of \model\ over them. We also presented a detailed ablation study to justify the use of each module in \model\ and presented error analysis in terms of confusion matrices and qualitative predictions to strengthen our results. We further highlighted the shortcomings of \model\ by highlighting examples from the test set of \dataset. While \model\ performed better than the comparative techniques, the performance attained might not be competitive for a real-world application. The fact that we had fewer persona instances in our curated dataset could be one of the contributing factors towards the scarcity of performance. As observed in Table \ref{tab:standalone_ph2} and Table \ref{tab-e2e}, the performance for \model\ boosted as the number of samples for the persona slot increased (maximum for \textit{likes} and minimum for \textit{occupation}). Leaning on the same reasoning, we posit that while the lack of competitive performance was a limitation for us, it could be mitigated by incorporating additional data. In the future, we would like to advance the model architecture to deal with the task efficiently. We will also focus on dynamically discovering the persona types rather than defining them in advance. The final goal would be to incorporate the discovered persona information into the dialogue generation model to generate meaningful and more engaging utterances.